# Automotive Crash Dynamics Modeling Accelerated with Machine Learning

**Mohammad Amin Nabian**
NVIDIA

**Sudeep Chavare**
General Motors

**Deepak Akhare , Rishikesh Ranade , Ram Cherukuri and Srinivas Tadepalli**
NVIDIA

## Abstract

Crashworthiness assessment is a critical aspect of automotive design, traditionally relying on high-fidelity finite element (FE) simulations that are computationally expensive and time-consuming. This work presents an exploratory comparative study on developing machine learning-based surrogate models for efficient prediction of structural deformation in crash scenarios using the NVIDIA PhysicsNeMo framework. Given the limited prior work applying machine learning to structural crash dynamics, the primary contribution lies in demonstrating the feasibility and engineering utility of the various modeling approaches explored in this work. We investigate two state-of-the-art neural network architectures for modeling crash dynamics: MeshGraphNet, a graph neural network that is widely employed in physics-based simulations, and Transolver, a transformer-based architecture with a physics-aware attention mechanism designed to maintain linear computational complexity with respect to geometric scale. Additionally, we examine three strategies for modeling transient dynamics: Time-Conditional, where the temporal state is directly parameterized by time; the standard Autoregressive approach, which recursively propagates predictions through time; and a stability-enhanced Autoregressive scheme incorporating rollout-based training to improve prediction accuracy and long-term temporal consistency. The models are evaluated on a comprehensive Body-in-White (BIW) crash dataset comprising 150 detailed FE simulations using LS-DYNA. The dataset represents a structurally rich vehicle assembly with over 200 components, including 38 key components featuring variable thickness distributions to capture realistic manufacturing variability. Each model utilizes the undeformed mesh geometry and component characteristics as inputs to predict the spatiotemporal evolution of the deformed mesh during the crash sequence. Evaluation results show that the models capture the overall deformation trends with reasonable fidelity, demonstrating the feasibility of applying machine learning to structural crash dynamics. Although not yet matching full FE accuracy, the models achieve orders-of-magnitude reductions in computational cost, enabling rapid design exploration and early-stage optimization in crashworthiness evaluation.

The code for this work is available at:

[https://github.com/NVIDIA/physicsnemo/tree/main/examples/structural_mechanics](https://github.com/NVIDIA/physicsnemo/tree/main/examples/structural_mechanics)

## Introduction

In the modern automotive industry, the assurance of vehicle safety is not merely a design consideration but a fundamental engineering imperative. Crashworthiness, defined as the ability of a vehicle's structure to protect its occupants during an impact, is a primary driver of the design and validation process. This focus is enforced by stringent government regulations and amplified by consumer safety rating programs, which have created a competitive landscape where occupant protection is a key market differentiator. The engineering challenge of crashworthiness is profoundly complex, extending beyond simple structural strength. It involves the orchestrated management of kinetic energy during a collision, where specialized components like crumple zones are meticulously designed to deform in a controlled manner, absorbing impact energy and decelerating the vehicle to mitigate forces transferred to the occupant survival space.

The design of these safety-critical systems is an inherently iterative process. Engineers must evaluate a multitude of design variants, exploring the effects of geometric modifications, material selections, and component thicknesses on key performance indicators such as structural deformation, occupant compartment intrusion, and vehicle deceleration profiles [1]. Each design choice represents a compromise between safety, weight, cost, and manufacturability, necessitating a high volume of evaluations to converge on an optimal solution [2, 3].

### *High-Fidelity Finite Element Simulations*

For several decades, the primary tool for virtual crashworthiness assessment has been high-fidelity Finite Element Analysis (FEA) [4]. FEA revolutionized automotive design by providing a virtual alternative to the prohibitively expensive and time-consuming process of building and destructively testing physical prototypes. The finite element method operates by discretizing a complex vehicle structure into a mesh of smaller, simpler elements. By solving the fundamental equations of motion and material behavior for each element, FEA can provide a detailed, full-field prediction of the dynamic, non-linear events that unfold during a crash, including large plastic deformations, buckling, and component interactions [5].

The historical development of FEA for crash simulation reflects a continuous pursuit of higher fidelity, driven by advances in computational power. Early, simplified approaches using spring-mass models gave way to sophisticated continuum models employing millions of shell and solid elements to represent a full vehicle with remarkable geometric detail. This evolution was critically enabled by the development of explicit time integration solvers, which are well-suited to handle the highly transient, non-linear dynamics characteristic of impact events, including complex contact and material folding phenomena that implicit solvers struggled with [4, 6].

Today, FEA stands as the undisputed gold standard for virtual vehicle validation.

### *Limitations of Traditional Methods*

While FEA was developed to overcome the bottleneck of physical testing, the relentless increase in vehicle complexity, material diversity, and safety requirements has ironically transformed high-fidelity FEA itself into a significant computational bottleneck in the modern design cycle. The very detail that makes FEA so powerful is also the source of its primary limitation: immense computational cost. A single, full-vehicle crash simulation can demand more than 15 hours of computation on a 16-CPU high-performance computing (HPC) cluster [7]. In the early days of its application, such simulations could take months to complete, rendering the results obsolete before they were even obtained.

This computational burden is particularly acute in the early stages of design, where the freedom to innovate is greatest and the need for rapid feedback is most critical. The cost of FEA scales poorly with model resolution; for some analyses, the computational complexity can be cubic in the number of nodes, meaning that a desire for higher accuracy through mesh refinement leads to an exponential increase in simulation time. This reality forces a constant and often suboptimal trade-off between fidelity and speed. Analysts must make expert judgments about where to apply a fine mesh (e.g., the front crumple zone) and where a coarse mesh will suffice (e.g., the rear of the vehicle), a process that requires significant manual effort and introduces potential inaccuracies. The high cost per simulation severely constrains the number of design alternatives that can be explored, stifling innovation and extending development timelines.

This dynamic illustrates a recurring pattern in engineering: the tool developed to solve one generation's bottleneck becomes the bottleneck for the next. Physical prototypes were once the limiting factor, leading to the development of FEA. Now, the complexity of modern design has pushed FEA into that same role, creating a clear and urgent need for the next evolution in simulation technology.

### *A New Paradigm: AI-Driven Surrogates for Accelerated Simulation*

In response to the computational challenge posed by FEA, a new paradigm is emerging: the use of AI-driven surrogate models. A surrogate model, also known as a metamodel or response surface, is a data-driven approximation of a complex, computationally expensive function or simulation. Instead of solving the governing physical equations from first principles for every new design, a surrogate model learns the underlying input-output mapping directly from data generated by a set of high-fidelity simulations. Once trained, these models can make predictions in milliseconds to seconds, offering a speed-up of several orders of magnitude compared to the original simulation [8].

The recent confluence of powerful deep learning algorithms and scalable computing hardware has catalyzed a revolution in this field. Neural network-based surrogates have demonstrated an extraordinary capacity to approximate the highly non-linear, high-dimensional physics that govern complex engineering systems, from fluid dynamics to structural mechanics. This transformative capability allows the engineering workflow to shift from conducting a few, carefully selected, time-consuming simulations to performing comprehensive, near-instantaneous exploration of the entire design space [9].

This shift represents more than a mere increase in speed; it is a fundamental change in the methodology of scientific computation, moving from a physics-first to a data-first approach. An FEA solver's "knowledge" is encoded in the mathematical formulation of physical laws, such as the conservation of momentum. In contrast, an AI surrogate's "knowledge" is encoded in the learned weights of a neural network, derived entirely from observing the results of those physical laws as manifested in simulation data. This establishes a new economic model for simulation: a significant upfront investment in generating a diverse training dataset is amortized over a virtually limitless number of subsequent, near-zero-cost predictions. This model is exceptionally well-suited for the early-stage design process, where rapid iteration and broad exploration are paramount. A Comparative Analysis of FE simulations and ML models for crashworthiness assessment is presented in Table 1.

Table 1: Comparative Analysis of Simulation Methodologies for Crashworthiness Assessment.

| **Metric** | **FE Simulation** | **ML Model** |
|---|---|---|
| **Computation Time** | Hours to Days (per simulation) | Hours to Days (training), Seconds (per inference) |
| **Hardware Requirement** | HPC Clusters | HPC Clusters (training), Single GPU Workstation (inference) |
| **Design Iteration Speed** | Slow; limits exploration to a few candidate designs | Rapid; enables large-scale, automated design space exploration and optimization |
| **Underlying Principle** | Physics-Based: Numerically solves PDEs. | Data-Driven: Learns the PDE solution from a simulation dataset. |
| **Physical Fidelity** | High (Considered Ground Truth for virtual testing) | Medium to High (Learns to replicate ground truth within engineering tolerances). |
| **Flexibility to New Physics or out-of-distribution samples** | High: Can be implemented directly into the solver. | Low: Requires retraining or fine-tuning on new data. |

Surrogate modeling of full-vehicle crash dynamics using machine learning remains a nascent but rapidly emerging research area. Li et al. [10] proposed Recurrent Graph U-Net (ReGUNet), a graph neural network surrogate tailored for crashworthiness prediction of vehicle panel components; this model embeds recurrence to track temporal evolution and reports less than 1% error on B-pillar intrusion metrics in a side-crash case study. Guennec et al. [11] benchmarked classical reduced-order surrogates against neural field models for crash simulation data, highlighting the trade-offs between model compactness and expressivity in capturing spatiotemporal deformations. Meanwhile, Thel et al. [12, 13] introduced the Finite Element Method Integrated Networks (FEMIN) framework, in which large regions of a crash-simulation mesh were replaced by neural networks, thereby accelerating simulations while retaining coupling with the remaining FEM domain. André et al. [14] explored a more modular approach by embedding a feedforward neural network surrogate for connector (joint) models—e.g. self-piercing rivets and flow-drill screws—in large-scale explicit crash simulations. In the structural-dynamics domain more broadly, Wen et al. [15] proposed a GNN + Temporal Convolutional Network (TCN) hybrid to forecast nonlinear responses of irregular components represented as graphs, enabling efficient spatiotemporal prediction across complex geometries. To our knowledge, the present work is the first to apply a fully data-driven surrogate to capture the complete vehicle-level crash dynamics (i.e. all structural parts interacting over time), rather than focusing solely on panels, joints, or localized subsystems.

## *Scope of Exploration and Key Contributions*

Building on the motivation outlined above — the need to overcome the computational bottleneck of high-fidelity FEA while retaining physical accuracy — this work takes a systematic approach to explore the applicability and promise of advanced ML models for automotive crash modeling, laying the groundwork for future developments in this rapidly evolving area. By investigating the strengths and limitations of state-of-the-art neural architectures and temporal schemes, this study aims to chart a path toward the development of practical, high-fidelity ML surrogates that can transform how crashworthiness is evaluated.

We frame this exploration around two central objectives:

- Capturing geometric and structural interactions — understanding how different parts of the vehicle structure interact under complex loading conditions.
- Predicting long-term dynamic behavior — maintaining causality, physical accuracy, and stability over the full course of a crash event.

Achieving these objectives requires targeted exploration of model architectures capable of representing structural physics and temporal schemes that ensure robust dynamic prediction.

### Capturing Geometric and Structural Interactions

We investigate two leading architectures designed for unstructured mesh data, representing complementary approaches to modeling crash dynamics:

**Transolver:** A novel attention-based model that approaches the problem from a different angle. Rather than operating directly on the mesh topology, Transolver learns a latent physical state representation of the system and performs attention-based updates within this learned space. This allows it to overcome the quadratic complexity of standard Transformers and scale to large problems with irregular domains. Transolver is designed to capture both local and global dependencies in the physical system, making it a promising candidate for high-resolution, large-scale crash simulations where both accuracy and scalability are critical [16, 17].

**MeshGraphNet**: A high-performance Graph Neural Network (GNN) designed for mesh-based physical simulation. GNNs operate directly on the connectivity structure of meshes, representing finite element nodes as graph nodes and edges as connectivity relationships. This grants them a powerful relational inductive bias, enabling them to learn localized physical interactions consistent with underlying mechanics. MeshGraphNet's architecture has demonstrated strong capabilities for accurately modeling complex structural interactions while maintaining computational efficiency [18, 19, 20].

### Predicting Long-Term Dynamic Behavior: Transient Schemes

Accurate transient prediction is a core requirement for crash modeling, as the simulation must faithfully reproduce the entire time history of the event while maintaining physical plausibility and stability. We systematically evaluate three distinct transient prediction schemes:

**Time-Conditional (Non-Autoregressive):** In this approach, the model is trained to predict the system state at an arbitrary time step directly from the initial state and time condition, without stepping sequentially through intermediate states.

**Autoregressive with One-Step Training (AR-OT):** Here, the model learns to predict the state at the next time step given only the current state. To generate the full sequence, predictions are rolled forward step-by-step.

**Autoregressive with Rollout Training (AR-RT):** Here, the stability challenge is explicitly addressed by training the model over multi-step rollouts rather than isolated one-step predictions. During training, the model predicts several or the entire steps ahead, with its own outputs fed back as inputs. This forces the model to learn to correct its own accumulated errors, improving stability and robustness for long-term prediction. AR-RT is computationally more intensive during training but offers superior fidelity in transient prediction.

These schemes represent different trade-offs between flexibility, stability, and computational cost, and comparing them is a core part of this study.

### Validation

We validate this framework on a full vehicle Body-in-White (BIW) dataset with variable component thicknesses. The results demonstrate that the model can learn the complex relationship between initial design parameters and the full-field, time-evolving response of the structure, including deformation.

### Implementation using NVIDIA PhysicsNeMo

The ML surrogate models presented in this work have been developed using the NVIDIA PhysicsNeMo [21], a highly optimized framework for building, training, and deploying physics-ML models. PhysicsNeMo provides an end-to-end infrastructure specifically designed for large-scale simulation-based learning tasks, making it well-suited for computationally demanding applications such as automotive crash modeling.

Unlike single-model repositories, PhysicsNeMo is a general-purpose framework that enables experimentation with a wide range of model architectures—such as graph neural networks, attention-based models, and hybrid formulations—within a unified environment. This design eliminates the need to stitch together multiple repositories or frameworks, significantly simplifying model development and comparison.

Key features of PhysicsNeMo that were leveraged in this study include:

- **Distributed Training:** PhysicsNeMo natively supports multi-GPU and multi-node training, allowing large models and high-resolution datasets to be processed efficiently. This capability was critical for training MeshGraphNet and Transolver architectures on full-vehicle Body-in-White datasets with hundreds of thousands of nodes.
- **Optimized Network Architectures:** The framework provides highly optimized implementations of both graph-based and attention-based neural architectures. These optimizations reduce computational overhead and improve training throughput.
- **Data Pipelines**: The framework offers flexible data pipelines for efficiently handling large simulation datasets, including support for variable-sized meshes, normalization, graph construction, and streamlines the workflow from raw finite element data to model-ready inputs.
- **Modular and Extensible Design**: PhysicsNeMo's modular architecture allows for easy integration of custom components, making it straightforward to experiment with novel architectures, transient schemes, or physics-informed loss functions.

By building on PhysicsNeMo, we were able to implement train the models on large-scale crash datasets efficiently and explore multiple transient prediction schemes. The framework's performance optimizations and distributed training capabilities were instrumental in enabling rapid experimentation and evaluation, laying the foundation for scalable, production-ready ML surrogates for crash dynamics.

# Methodology – Model Architectures

## *Transolver*

The standard Transformer architecture, which uses a self-attention mechanism to compute interactions between all pairs of input tokens, is computationally prohibitive for typical FE models, which can contain hundreds of thousands or millions of nodes. The complexity of canonical attention is quadratic with respect to the number of nodes, making it infeasible for large-scale industrial simulations.

The Transolver architecture [16, 17] overcomes this limitation by operating on a more foundational idea: learning the intrinsic physical states hidden within the discretized geometry. The core innovation is a novel Physics-Attention mechanism that first decomposes the computational domain into a small, learnable set of "slices" representing distinct physical states. It then applies the powerful attention mechanism to a compressed representation of these slices, known as "physics-aware tokens." This reduces the problem from attending over $N$ nodes to attending over $M$ tokens, where $M \ll N$, achieving linear complexity with respect to the number of mesh points and enabling scalability to massive industrial problems.

This methodology can be understood as a form of learned, soft-probabilistic domain decomposition. Where engineers might manually partition a complex geometry into regions for parallel computation, Transolver learns to identify and group regions that behave in a physically similar manner—for example, areas of high plastic strain, regions undergoing rigid body motion, or zones experiencing shock wave propagation. By first learning an efficient representation of the problem domain itself, the model can then focus its capacity on learning the complex interactions between these physically meaningful regions.

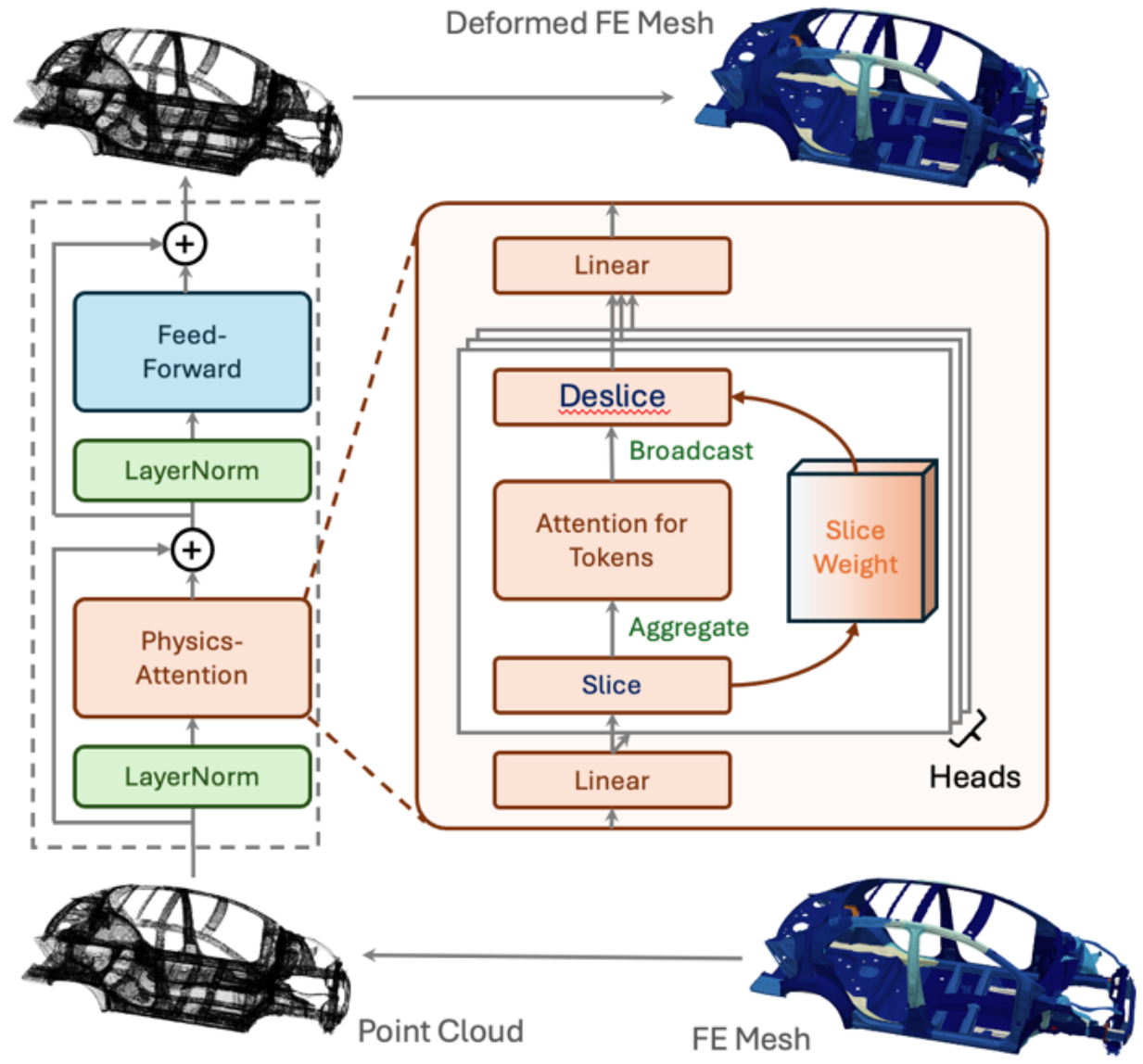

Figure 1. The transolver architecture.

### The Physics-Attention Mechanism

The Physics-Attention mechanism is the core building block of the Transolver model. It replaces the standard self-attention layer in a Transformer and can be broken down into a sequence of four key operations: slice weight generation, physics-aware token encoding, attention on tokens, and full-field reconstruction via deslicing.

The process begins by treating the FE mesh as a point cloud. Each of the $N$ mesh points, described by a feature vector $x_i \in R^c$ (encoding geometric and physical information like coordinates, material properties, etc.), is processed to determine its affiliation with $M$ learnable "slices." Each input feature vector $x_i$ is first passed through an initial linear layer to embed it into a unified feature space. The embedded feature of each mesh point is then projected into an M-dimensional vector using a learnable linear layer. A Softmax activation is applied to this vector to produce a set of slice weights, $w_i \in R^M$:

$$\{w_i\}_{i=1}^{N} = S\big(P(x_i)\big), \qquad (1)$$

where $S$ is a Softmax function and $P$ is a projection layer. Each component $w_{i,j}$ of the vector $w_i$ represents the probabilistic degree to which mesh point $i$ belongs to slice $j$. The Softmax function ensures that $\sum_{j=1}^{M} w_{i,j} = 1$ for each point and encourages low-entropy assignments, meaning the model learns to assign points to a small number of slices with high confidence, promoting the emergence of distinct and informative physical states.

With the slice weights computed for every point, the next step is to create a compressed representation of the entire system. This is achieved by generating $M$ "physics-aware tokens," where each token represents the aggregated state of its corresponding slice. The j-th token, $Z_j \in \mathrm{R}^{c}$, is calculated as a normalized, weighted sum of all mesh point features:

$$z_j = \frac{\sum_{i=1}^{N} w_{i,j} x_i}{\sum_{i=1}^{N} w_{i,j}}. \quad (2)$$

This operation is a critical dimensionality reduction step. The entire state of the system, originally described by $N$ high-dimensional vectors, is now compactly represented by just $M$ physics-aware tokens. Crucially, because this aggregation is a sum over all points, the token generation process is permutation invariant. The order in which the mesh points are processed does not affect the final token representation, which is a key property for handling unstructured data and generalizing across different mesh discretizations.

Once the system is represented by the sequence of M tokens, a standard multi-head self-attention mechanism can be applied. This allows the model to learn the intricate, long-range correlations and interactions between the different learned physical states.

**Attention**: The tokens $\{z_j\}_{j=1}^{M}$ are used to compute query ($q$), key ($k$), and value ($v$) vectors via linear projections. The attention mechanism then computes the updated token representations $\{z_j'\}_{j=1}^{M}$:

$$q, k, v = \mathrm{Linear}(z), \quad (3)$$

$$z' = \mathrm{Softmax}(C\, qk^{\mathrm{T}})v. \quad (4)$$

**Field Reconstruction (Deslicing)**: After the attention mechanism has updated the tokens, the final step is to map this information back to the original mesh points to produce a full-field prediction. This "deslicing" operation reconstructs the updated feature vector for each mesh point, $x_i'$, as a weighted sum of the updated tokens, using the original slice weights $w_{i,j}$:

$$x_i' = \sum_{j=1}^{M} w_{i,j} z_j'. \quad (5)$$

This process effectively broadcasts the information learned at the latent-state level back to the full spatial domain, producing the final output for that layer.

## The Transolver Architecture

The overall Transolver architecture consists of a stack of these Physics-Attention layers. Each layer is typically followed by a feed-forward network (MLP) and layer normalization, mirroring the design of a standard Transformer block. By stacking these layers, the model can iteratively refine its understanding of the physical states and their interactions, leading to a highly accurate approximation of the PDE's solution operator. The model's independence from mesh connectivity and its permutation-invariant nature are the sources of its claimed "endogenetic geometry-general modeling capacity," suggesting a powerful potential to generalize not just to unseen parameters but to unseen geometries and mesh topologies.

## *MeshGraphNet*

MeshGraphNet (MGN) [18, 19, 20] is a graph-based neural network designed to simulate physical systems by directly representing a mesh structure as a graph. The core idea is to leverage message passing between nodes to propagate physical state information — such as velocity, pressure, or temperature — across the mesh over time. This paradigm enables MGN to capture both local and global interactions in systems governed by partial differential equations (PDEs).

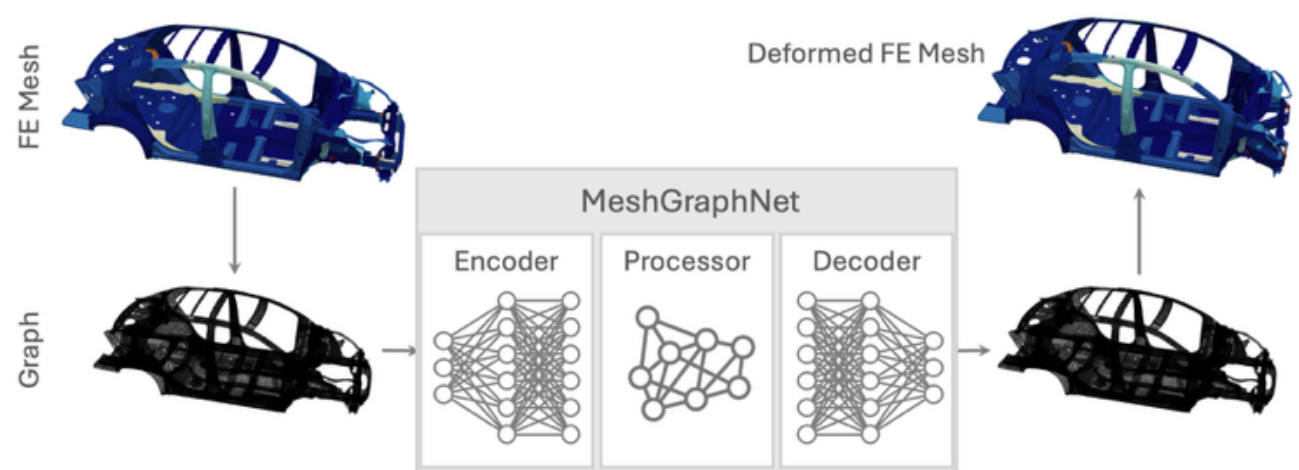

Figure 2. The MeshGraphNet architecture.

In MeshGraphNet, a mesh is represented as a graph

$$G = (V, E), \quad (6)$$

where $V$ is the set of nodes, corresponding to the mesh vertices, $E$ is the set of edges, corresponding to connections between adjacent vertices. Each node $i \in V$ is associated with a feature vector

$$h_i \in \mathbb{R}^d, \quad (7)$$

which encodes relevant physical quantities such as velocity $v_i$, pressure $p_i$, or temperature. Each edge $(i, j) \in E$ has a feature vector

$$e_{ij} \in \mathbb{R}^k, \quad (8)$$

which encodes information about the relationship between nodes $i$ and $j$, such as relative position or distance.

The MGN follows an Encode–Process–Decode architecture as shown in Fig 2. In this framework, an encoder network first maps the raw node and edge features (e.g., physical quantities and geometric attributes) into a latent space through independent MLPs. This encoding step ensures that heterogeneous input features are projected into a unified, high-dimensional representation suitable for message passing. The processor, implemented as a sequence of message-passing layers, iteratively refines these latent representations by exchanging information between connected nodes and edges, allowing the network to learn complex spatial dependencies and nonlinear interactions. Finally, a decoder network transforms the processed latent node features back into the target physical quantities—such as accelerations, displacements, or fluxes.

**Message Passing Mechanism**

The central mechanism of MeshGraphNet is message passing, which enables iterative propagation of information between neighboring nodes. The process involves three main steps: message computation**,** message aggregation**,** and node update**.**

For each edge $(i, j)$, the model computes a message $m_{ij}$ from node $j$ to node $i$ as:

$$m_{ij} = \phi_m(h_i, h_j, e_{ij}), \tag{9}$$

where $\phi_m$is a learnable neural network function that produces $m_{ij} \in \mathbb{R}^d$.

The incoming messages to node $i$from its neighbors are aggregated. A common choice is summation:

$$m_i = \sum_{j \in \mathcal{N}(i)} m_{ij}, \tag{10}$$

where $\mathcal{N}(i)$denotes the set of neighbors of node $i$. The aggregated message $m_i$ contains information from the local neighborhood.

The node's features are updated based on the aggregated message:

$$h_i' = \phi_u(h_i, m_i), \tag{11}$$

where $\phi_u$ is a learnable neural network function, and $h_i'$ is the updated feature vector for node $i$.

The above message-passing process is repeated over $L$ layers of the graph neural network, each with distinct learnable parameters. With each layer, information propagates further through the graph, enabling nodes to capture interactions from larger neighborhoods.

The update rule at layer $l$can be expressed as:

$$h_i^{(l+1)} = \phi_u^{(l)}\left(h_i^{(l)}, \sum_{j \in \mathcal{N}(i)} \phi_m^{(l)}\left(h_i^{(l)}, h_j^{(l)}, e_{ij}\right)\right), \tag{12}$$

where $\phi_m^{(l)}$ and $\phi_u^{(l)}$ are learnable functions at layer $l$, and $h_i^{(l)}$ is the feature vector of node $i$ at layer $l$.

After $L$ layers of message passing, each node aggregates information from all nodes within its $L$-hop neighborhood, enabling the model to capture both local and global system dynamics.

Through this architecture, MeshGraphNet efficiently learns a physically consistent update rule for node states, enabling accurate and scalable simulation of complex physical systems such as automotive crash events.

**Multi-scale graph**

Training the MGN surrogate on a fine mesh with a large number of nodes and edges is computationally expensive. In such high-resolution meshes, each node's receptive field depends on the number of message-passing layers, and a finer discretization limits the spatial range of information exchange. To address both the computational cost and limited receptive field, a multi-scale graph is constructed inspired from the X-MeshGraphNet model [19].

First, a point cloud is extracted from the FE mesh nodes, and Farthest Point Sampling (FPS) is applied to subsample a set of nodes $V_s$ that are well separated spatially. Unlike random sampling, FPS selects nodes directly from the original point cloud rather than generating new ones, ensuring that all sampled nodes coincide with actual mesh locations. Next, the k-Nearest Neighbors (KNN) algorithm is used to identify local neighbors and construct edges $E_s$. o further enhances long-range connectivity, an additional subset of nodes is obtained using FPS, and KNN is applied again to form extended edges $E_{ss}$ connecting distant nodes. The resulting multi-scale graph denoted as $G_s = (V_s, \{E_s, E_{ss}\})$, provides a structure that reduces computational overhead while increasing each node's effective receptive field, thereby improving model efficiency and stability during training. The training can then be conducted on subsampled multiscale graph, a schematic of which can be found in Figure 3.

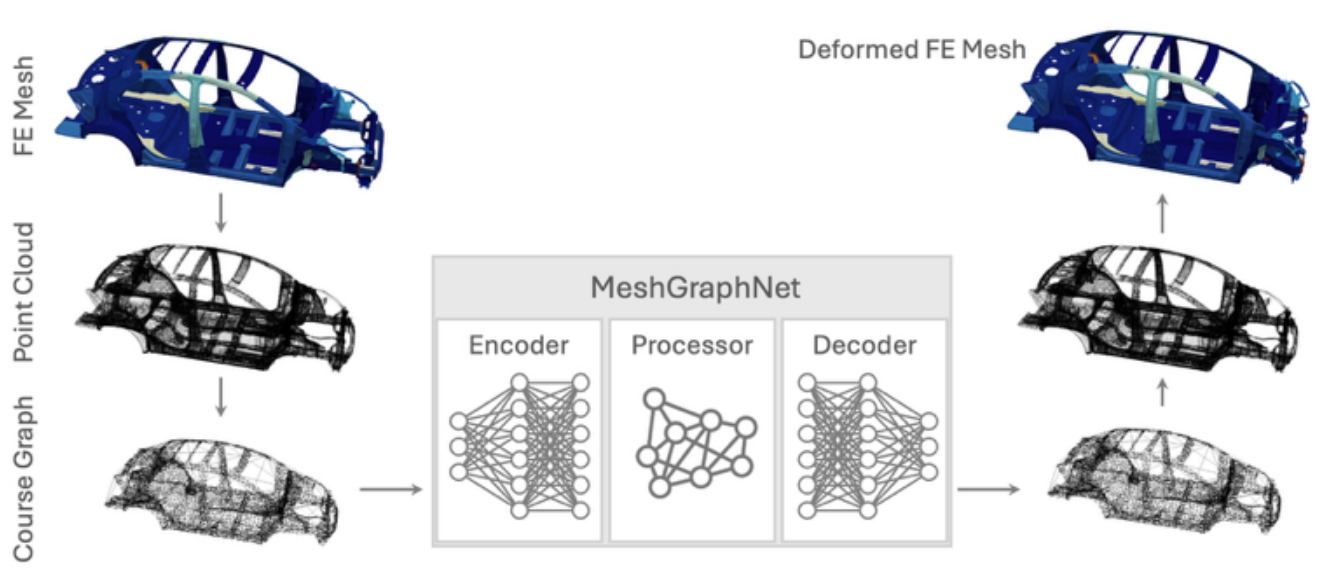

Figure 3. MeshGraphNet with multi-scale graphs.

# Methodology - Transient Dynamics Schemes

This study investigates three distinct training schemes, each embodying a different approach to modeling time evolution. These schemes differ in how they handle causality, error accumulation, and stability over long prediction horizons. The dataset can be represented as

$$D = \left\{\tau, \left\{X_{t_i}^{GT}\right\}_{i=0}^{T}\right\}_{m=0}^{N}, \tag{13}$$

where, $\tau$ is the thickness, $X_{t_i}^{GT}$is the state variable at time $t_i$, $T$ is the number of time steps, and $N$ is the number of samples. The $X_{t_i}^{GT} \in R^{N \times 3}$ is vector representing a point cloud with $N$ nodes that coincides with the nodes of the FE mesh. The thickness is provided at each node, therefore $\tau \in R^N$.

## *Scheme A - Time-Conditional (Non-Autoregressive)*

In the Time-Conditional (TC) scheme, the model directly predicts the system state at any given time step $t_i$ from the initial condition and the time itself:

$$X_t = \mathcal{M}_\theta(X_0^{GT}, t). \tag{14}$$

The dataset is modified as $D = \{(\tau, X_{t_0}^{GT}, t_i), X_{t_i}^{GT}\}_{m=0}^{N\times L}$ and the training objective minimizes the instantaneous prediction error independently for each time step:

$$\mathcal{L}_{TC} = \sum_{D} \| X_{t_i}^{GT} - \mathcal{M}_\theta(X_{t_0}^{GT}, t_i) \|^2, \tag{15}$$

where $X_{t_i}^{GT}$ is the ground truth state at time $t_i$, $X_{t_0}^{GT}$ is the initial condition, and $\mathcal{M}_\theta$ denotes the model.

The model is trained to predict the state at any arbitrary time, thereby capturing continuous-time dynamics without explicit temporal recursion. The main advantage of this approach is computational efficiency and parallelism, since each time step can be trained independently and avoids accumulation of autoregressive errors since there is no stepwise feedback. However, it fundamentally ignores causal dependencies between states, and thus tends to perform poorly when extrapolating or forecasting beyond the training time horizon.

### *Scheme B - Autoregressive with One-Step Training*

The Autoregressive scheme with one-step training (AR-OT) focuses on learning the transition function:

$$X_{t_{i+1}} = F_\theta(X_{t_i}^{GT}), \tag{16}$$

where $F_\theta$ is the learned transition operator.

During training, the current state is taken directly from the ground truth data — a technique known as *Teacher Forcing*. So, the dataset for training becomes $D = \{(\tau\, X_{t_i}^{GT}), X_{t_{i+1}}^{GT}\}_{m=0}^{N\times L}$ and the objective to minimize the one-step prediction error is given as:

$$\mathcal{L}_{AR} = \sum_{D} \| X_{t_{i+1}}^{GT} - F_\theta(X_{t_i}^{GT}) \|^2. \tag{17}$$

The model learns to predict the next state given the exact ground truth state from the previous time step. This approach ensures high per-step accuracy and stability during training. However, it suffers from **covariate shift**: at inference time, predicted states - rather than ground truth states - must be used as inputs. This leads to error accumulation, especially in highly non-linear systems such as crash dynamics, where small deviations compound rapidly. Without explicit training to handle these deviations, the model's rollout performance degrades significantly over long sequences.

### *Scheme C — Autoregressive with Rollout Training*

Autoregressive with Rollout Training (AR-RT) explicitly aligns the training objective with the inference scenario by training over multi-step trajectories rather than single-step predictions. The model learns the transition function

$$X_{t_{i+1}} = F_\theta(X_{t_i}), \tag{18}$$

by minimizing the loss computed over a rollout of length $L$:

$$\mathcal{L}_{AWRT} = \sum_{m=1}^{N} \sum_{k=1}^{L} \| X_{t_k}^{GT} - X_{t_k} \|^2, \tag{19}$$

where $X_{t_k}^{GT}$ is the ground truth and $X_{t_k}$ is the model prediction obtained recursively as $X_{t_{i+1}} = F_\theta \cdot F_\theta \cdots F_\theta(X_{t_0}^{GT})$.

Training involves predicting multiple time steps ahead $\{X_{t_i}\}_{i=1}^{L}$ for given $\tau$ and $X_{t_0}$, with the model's own predictions fed back as inputs for subsequent steps. This explicitly trains the model to handle propagated errors. To enforce stability and robustness, gradients must flow through the entire rollout sequence, analogous to Backpropagation Through Time (BPTT) in recurrent networks. This ensures that the model learns a robust transition function $F_\theta$ that remains effective even under imperfect or perturbed inputs, thereby minimizing long-term error accumulation rather than optimizing solely for one-step accuracy. AR-RT thereby instills a strong temporal inductive bias.

To ensure numerical stability during training, the MGN and Transolver networks -- denoted as $f_\theta(\cdot)$ -- are formulated to predict the nodal accelerations $\ddot{X}_{t_i}$ rather than the states directly. The temporal evolution of the system is then reconstructed using standard ODE time-integration schemes, which update the nodal positions and velocities at each time step according to the predicted accelerations, as follows:

$$\dot{X}_{t_{i+\frac{1}{2}}} = \Delta t \cdot \ddot{X}_{t_i} + \dot{X}_{t_{i-\frac{1}{2}}}, \qquad X_{t_{i+1}} = \Delta t \cdot \dot{X}_{t_{i+\frac{1}{2}}} + X_{t_i}, \tag{20}$$

where $\dot{X}_{t_{i-\frac{1}{2}}} = \frac{X_{t_i} - X_{t_{i-1}}}{\Delta t}$. Therefore, the transition function is

$$F_\theta(X_{t_i}, X_{t_{i-1}}, \tau) = \Delta t^2 \cdot f_\theta(X_{t_i}, X_{t_{i-1}}, \tau) + 2X_{t_i} - X_{t_{i-1}}. \tag{21}$$

## Case Study: Predicting Deformation in BIW Crash

### *The Body-in-White (BIW) Dataset*

The dataset is derived from a high-fidelity finite element model (FEM) of a Body-in-White (BIW) structure, simplified from a public-domain model provided by the National Highway Traffic Safety Administration [22]. The BIW model contains approximately 400,000 nodes and 380,000 elements, preserving key crash-relevant structural characteristics while reducing computational complexity.

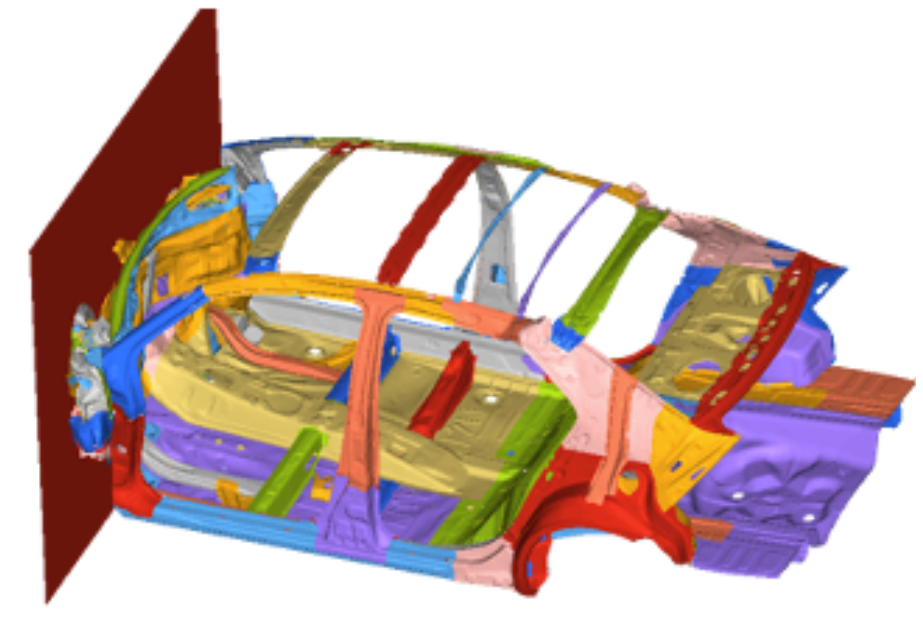

Figure 4. Simplified vehicle body used for crash simulation.

Crash simulations were performed using Finite Element Analysis (FEA) within LS-DYNA, a solver well suited for nonlinear dynamic problems involving large deformations, contact interactions, and complex material behavior. The vehicle was impacted against a rigid barrier under a 56 kph frontal crash scenario, consistent with standardized test configurations. The model includes essential structural features—crumple zones, reinforcements, and energy-absorbing members—that capture realistic crash deformation patterns.

To balance detail and efficiency, the vehicle body subsystem was used in place of the full-vehicle model, with a 10 mm mesh size selected to ensure sufficient resolution for localized deformations. Each simulation was run for 120 milliseconds and completed in approximately 10 minutes on General Motors' HPC cluster.

A Design of Experiments (DoE) was conducted by varying the thickness of 33 front-end components within ±20% of their nominal values (Figure 5). This yielded 150 unique designs. For each design, LS-DYNA simulations were executed, providing high-fidelity simulations to capture detailed deformation, acceleration responses, enabling robust data-driven modeling of vehicle crash dynamics and structural performance. The resulting data is partitioned into training (90%), validation (5%), and testing (5%) sets.

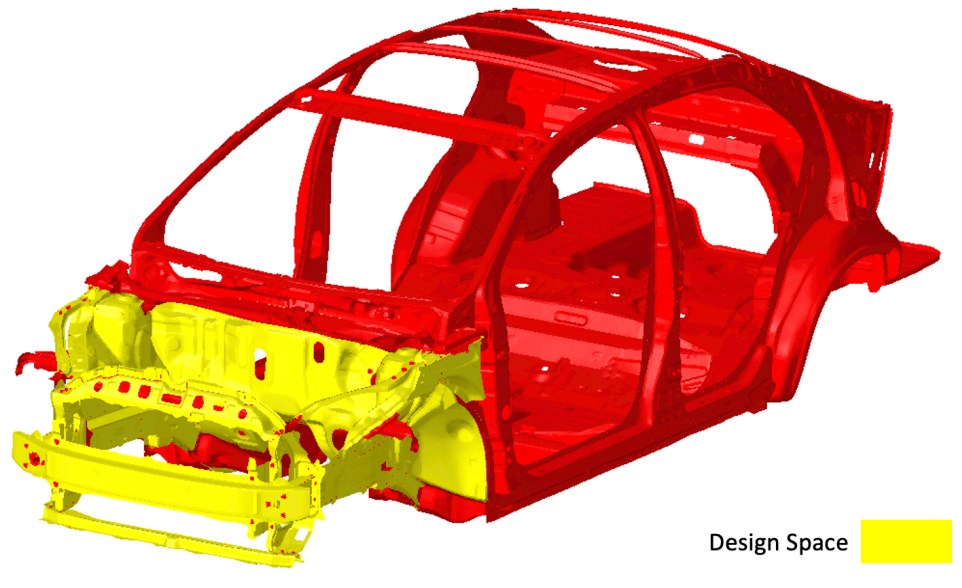

Figure 5. Parts included in the design space.

Figure 6 shows the probe points located at the driver and passenger toe pans, where acceleration responses were recorded during each crash simulation. These points were selected because the toe pan region is highly sensitive to front-structure deformation and provides a reliable indicator of the vehicle's structural integrity and occupant safety performance during frontal impacts.

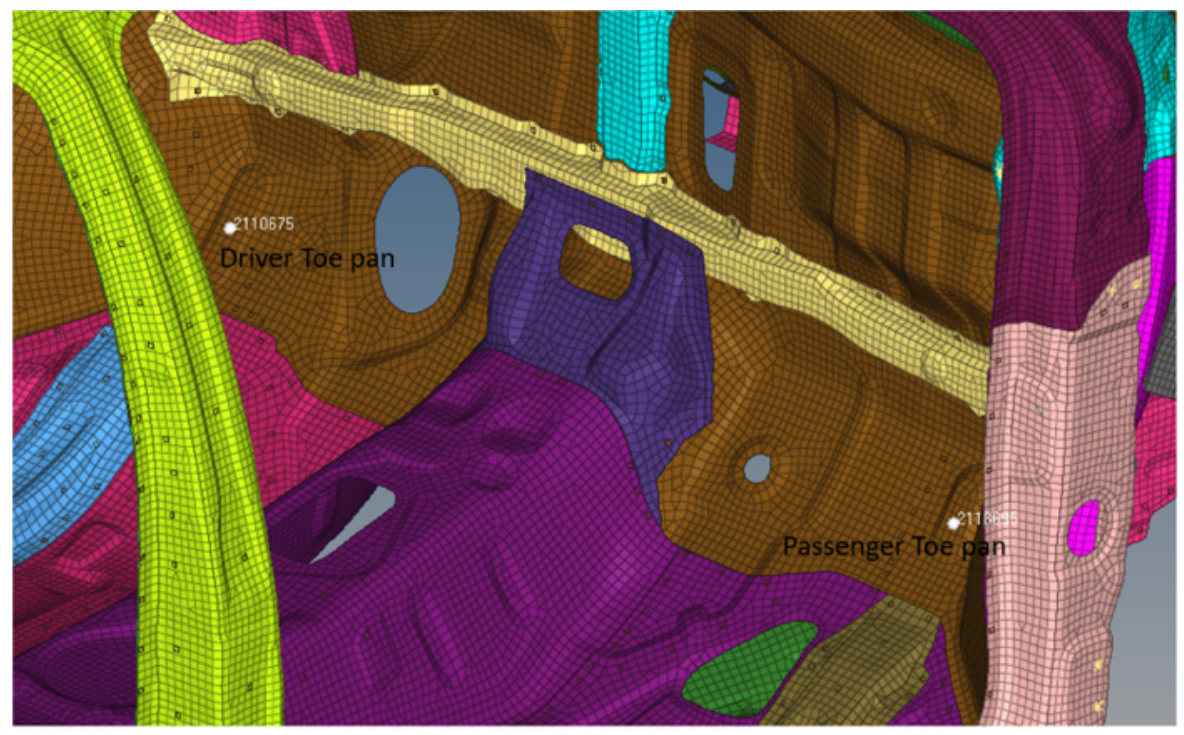

Figure 6. Probe points to measure the acceleration at the driver and passenger Toe Pans.

## *Results*

This section is organized into three parts. In the first part, we present a comprehensive evaluation of the Transolver architecture trained using the autoregressive rollout (AR-RT) scheme for crash surrogate modeling. This includes quantitative and qualitative analyses of its ability to capture spatiotemporal crash dynamics, assessing deformation accuracy, stability over time, and the accuracy of displacement, velocity and acceleration predictions at key probe points.

In the second part, we compare the performance of Transolver with MeshGraphNet under equivalent transient modeling conditions. This comparison highlights the strengths and limitations of each architecture in terms of accuracy, stability, and computational efficiency for predicting complex crash behavior.

In the third part, we examine the effects of different transient modeling schemes—Time-Conditional, Autoregressive with One-Step Training (AR-OT), and Autoregressive Rollout Training (AR-RT)—within the Transolver framework. This comparison investigates how the choice of transient scheme influences prediction accuracy, long-term stability, and the ability to capture detailed deformation patterns over the course of a crash event.

Due to the significant computational overhead of the rollout training scheme, which requires backpropagation through time, we use gradient checkpointing. Specifically, our approach checkpoints each timestep during rollout, substantially reducing memory usage required for training. With this strategy, we can perform training and inference across the full set of dataset timesteps. However, to make computation tractable across the range of models and approaches presented in this study, we limit the number of timesteps to 14, rather than the full 21 used in the dataset.

Both the Transolver and MeshGraphNet (MGN) models use a cosine annealing learning rate schedule, decaying from $10^{-4}$ to $10^{-6}$, and are trained for 8,000 epochs using mixed-precision (AMP) training for computational efficiency. MGN employs the ReLU activation function, whereas Transolver uses GELU.

For Transolver, we use 128 latent slices, 6 layers, a hidden dimension of 256, and 8 attention heads. For MGN, the model consists of 15 processor layers with sum aggregation, a hidden dimension of 128, and two MLP layers per message-passing operation in the processor. The encoder and decoder networks each use two MLP layers as well. The computational time per epoch for both Transolver and MGN models (using the original mesh) was around 110 seconds using a batch size of 1 on 8 H100 GPUs.

### Part 1 — Transolver with Autoregressive Rollout Training

We begin by evaluating the performance of the Transolver architecture trained using the autoregressive rollout (AR-RT) scheme, which was specifically designed to improve long-term prediction stability by explicitly training over multi-step trajectories. This approach is particularly relevant for crash dynamics, where small errors can rapidly accumulate over time.

Figure 7 illustrates the deformation predictions of Transolver-AR-RT for Sample #4 in the test dataset. Across all four viewing angles, the predicted deformed mesh closely aligns with the ground truth, with differences concentrated in regions of high deformation gradient. The displacement magnitude differences remain localized and small, confirming the model's ability to preserve physical fidelity over the full rollout.

Figure 8 further quantifies this performance by comparing displacement, velocity, and acceleration time histories at the driver and passenger toe pans. Transolver-AR-RT maintains consistent alignment with ground truth trajectories for displacement throughout the entire simulated crash sequence. However, we acknowledge that the accuracy of velocity predictions — and more significantly, acceleration predictions — remains suboptimal. We hypothesize that this limitation arises from the current loss function, which considers only the deviations between predicted and ground truth displacements. Incorporating additional loss terms that explicitly account for velocity and acceleration deviations may improve prediction accuracy for these quantities and represents a promising direction for future work.

Figures 9 and 10 present analogous results for Sample #103 in the test dataset, and additional sample results are provided in the Appendix. Figure 11 illustrates the cumulative displacement error growth over time for Transolver-AR-RT.

### Part 2 — MeshGraphNet with Autoregressive Rollout Training

Given the strong capability of MGN in capturing local spatial features, an MGN model was trained using the same autoregressive rollout (AR-RT) scheme for comparison. The MGN was trained on a graph constructed from a mesh containing 84,862 nodes and 294,6062 edges. To further improve computational efficiency, a Multiscale MGN was also investigated. A reduced-resolution graph was constructed containing approximately one-tenth the number of nodes in the original mesh. This modification reduced the computational time per epoch to 16 seconds.

Figure 12 provides a comparison of the frontal deformed mesh between MGN with original mesh, MGN with multiscale mesh, Transolver, and the ground truth for Sample #4 in the test dataset. All the models use the AR-RT scheme. A slight spatial noise is visible in the MGN predictions compared to Transolver-AR-RT. The Multiscale MGN predictions exhibit more spatial noise than those of the other models. However, this is due to the interpolation error, and it is hypothesized that this noise can be mitigated through a corrector or super-resolution neural network, which may refine the coarse predictions and recover high-fidelity structural details.

Figure 13 shows the cumulative displacement error growth over time for MGN-AR-RT, aggregated over the test samples. The error is noticeably higher than that of Transolver-AR-RT, indicating the superior predictive accuracy of the Transolver architecture. As shown in Figure 14, the cumulative displacement error over time for Multiscale MGN-AR-RT remains comparable to that of the original MGN-AR-RT, demonstrating that similar predictive performance can be achieved at significantly lower computational cost.

Figure 15 compares the displacement, velocity, and acceleration histories at the driver and passenger toe pans. Similar to Transolver-AR-RT, Multiscale MGN-AR-RT maintains close alignment with the ground-truth displacement trajectories throughout the crash sequence and performs suboptimally in capturing the finer variations in velocity and acceleration.

### Part 3 — Transolver with Other Transient Schemes

To further examine the impact of transient modeling strategies on predictive performance, we train and evaluate Transolver under three distinct temporal formulations: Time-Conditional, Autoregressive with One-Step Training (AR-OT), and Autoregressive Rollout Training (AR-RT). These formulations differ in how temporal dependencies are learned and propagated during both training and inference.

Figure 16 compares the Transolver predictions obtained using the time-conditional, AR-OT, and AR-RT schemes for Sample #4 in the test dataset. The relative L2 position errors for the time-conditional and AR-OT schemes are presented in Figures 17 and 18, respectively. The AR-OT scheme yields noticeably higher L2 position errors compared to AR-RT, whereas the time-conditional model achieves comparable accuracy with smaller standard deviation. Figures 19 and 20 further compare the displacement, velocity, and acceleration histories at the driver and passenger toe pans for the time-conditional and AR-OT schemes, respectively.

## *Analysis*

The results presented in this study demonstrate that both the Transolver and MGN architectures offer promising pathways for developing machine learning–based surrogate models for crash dynamics. Each model successfully captures the essential spatiotemporal evolution of structural deformation and exhibits relatively good agreement with high-fidelity FE simulations.

Overall, the Transolver architecture yields slightly higher predictive accuracy across all evaluated metrics, particularly in long-term deformation stability and displacement prediction. Its autoregressive rollout training (AR-RT) formulation proves especially effective in maintaining temporal consistency and preventing the accumulation of drift errors over extended trajectories. The transformer-based latent representation in Transolver allows for global context aggregation across the spatial domain, enabling it to capture large-scale deformation patterns more coherently than purely message-passing architectures.

In contrast, MGN remains a competitive and interpretable baseline, excelling at modeling local interactions within the mesh topology. Although its predictions display minor spatial noise relative to Transolver, the overall deformation trends remain physically consistent. The multiscale MGN variant demonstrates a particularly attractive trade-off between computational efficiency and accuracy. By reducing the number of nodes by roughly an order of magnitude, the training time per epoch decreases from approximately 110 seconds to just 16 seconds, while maintaining comparable displacement accuracy to the original MGN model. This highlights the potential of hierarchical or coarse-to-fine graph representations for scaling surrogate crash models to larger or more complex structures.

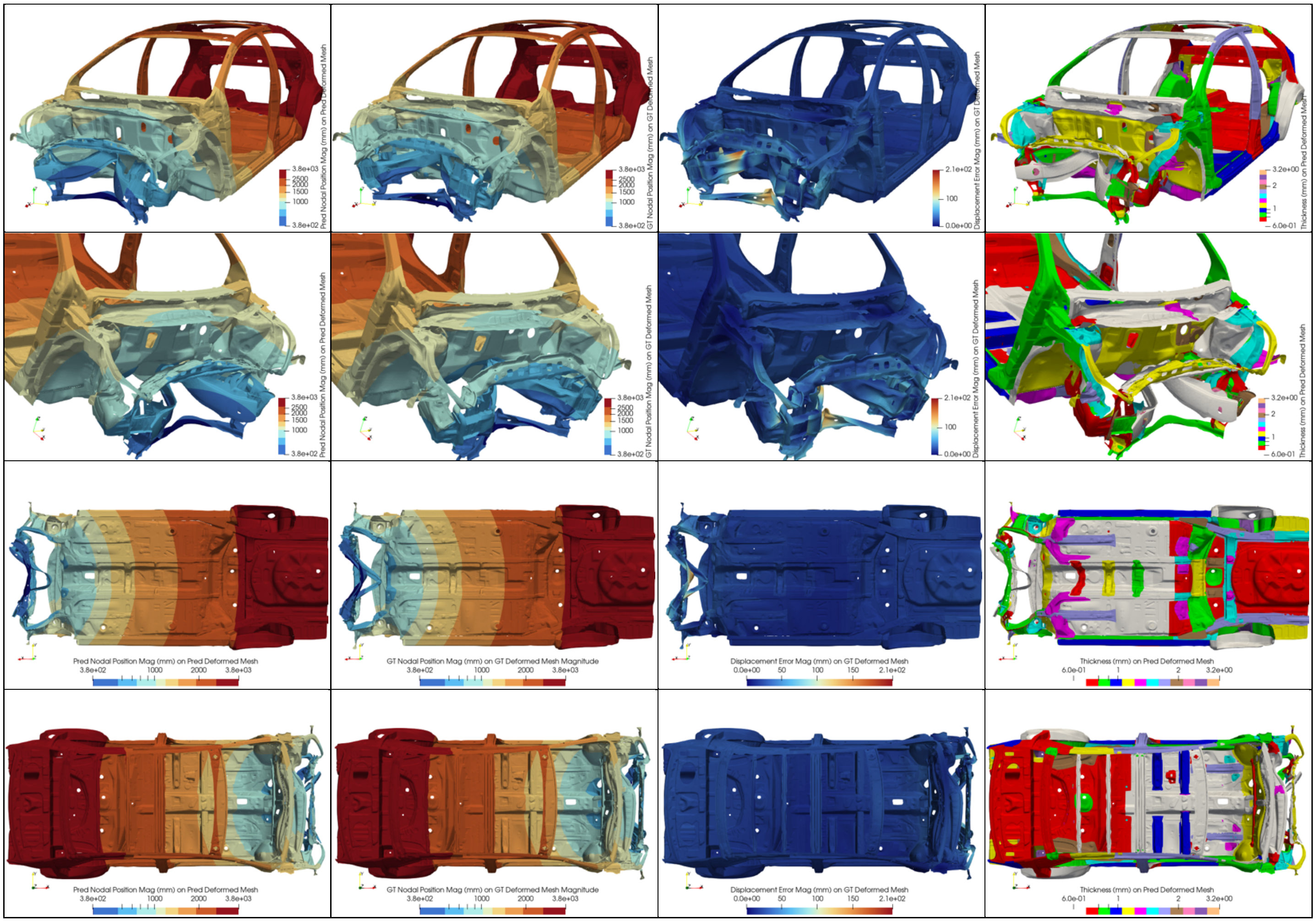

Figure 7. [Model: Transolver-AR-RT] Visualization of crash deformation results across four viewing angles (rows) for Sample #4 in the test dataset. Each column represents: (1) predicted deformed mesh colored by predicted nodal position magnitude, (2) ground-truth deformed mesh colored by ground truth nodal position magnitude, (3) displacement magnitude difference between prediction and ground truth plotted on the ground truth deformed mesh, and (4) component thickness plotted on the predicted deformed mesh.

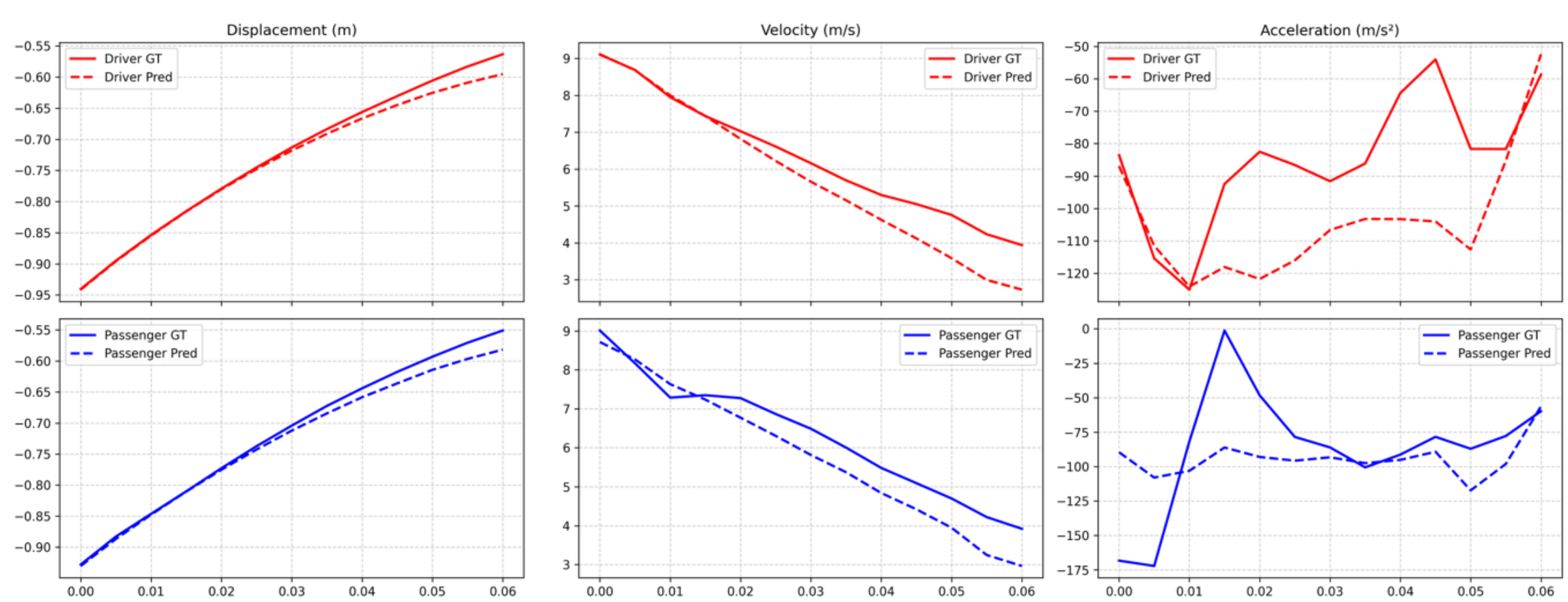

Figure 8. [Model: Transolver-AR-RT] Comparison between the predicted and ground-truth displacement, velocity, and acceleration at the driver and passenger toe pans for Sample #4 in the test dataset.

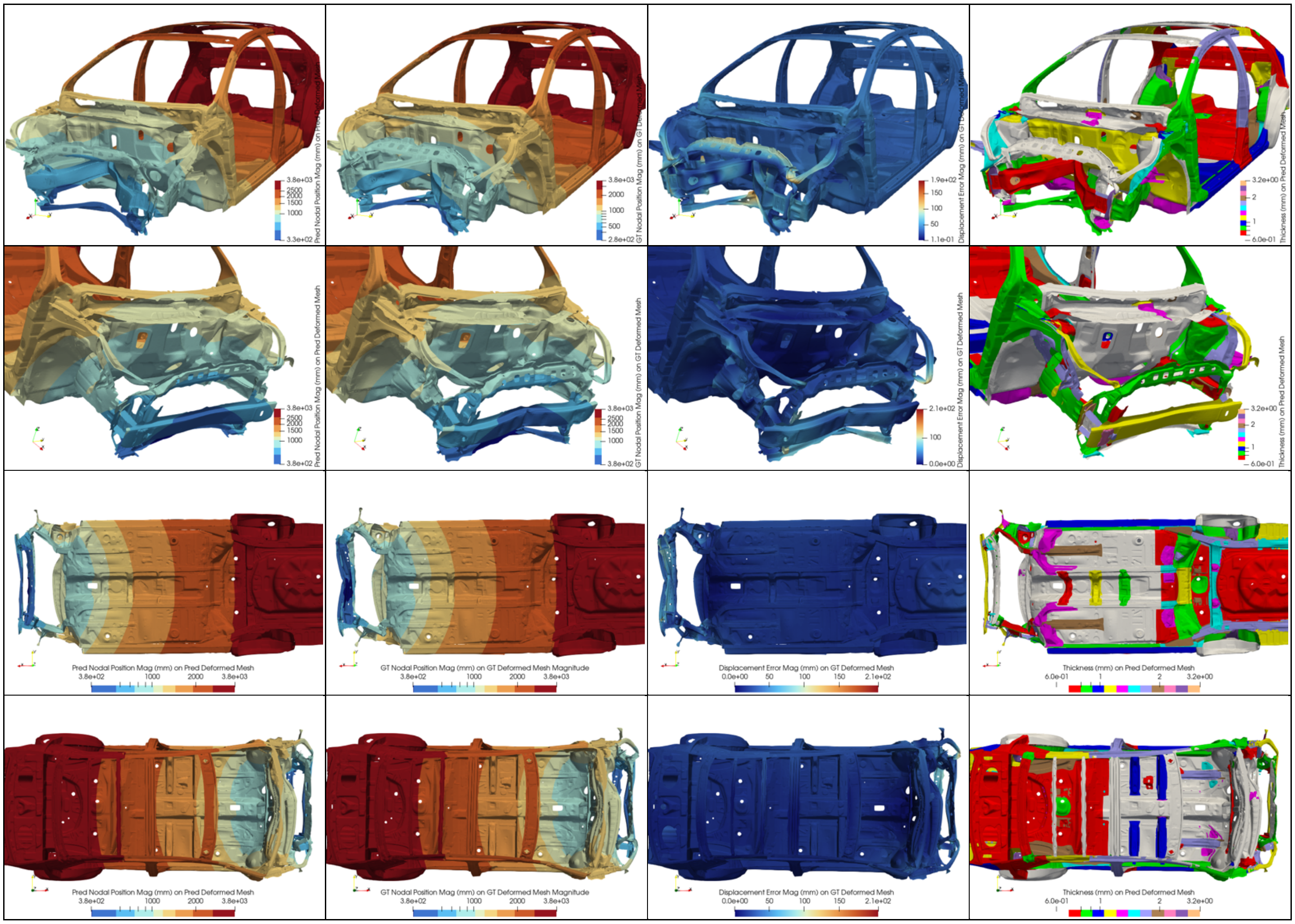

Figure 9. [Model: Transolver-AR-RT] Visualization of crash deformation results across four viewing angles (rows) for Sample #103 in the test dataset. Each column represents: (1) predicted deformed mesh colored by predicted nodal position magnitude, (2) ground-truth deformed mesh colored by ground truth nodal position magnitude, (3) displacement magnitude difference between prediction and ground truth plotted on the ground truth deformed mesh, and (4) component thickness plotted on the predicted deformed mesh.

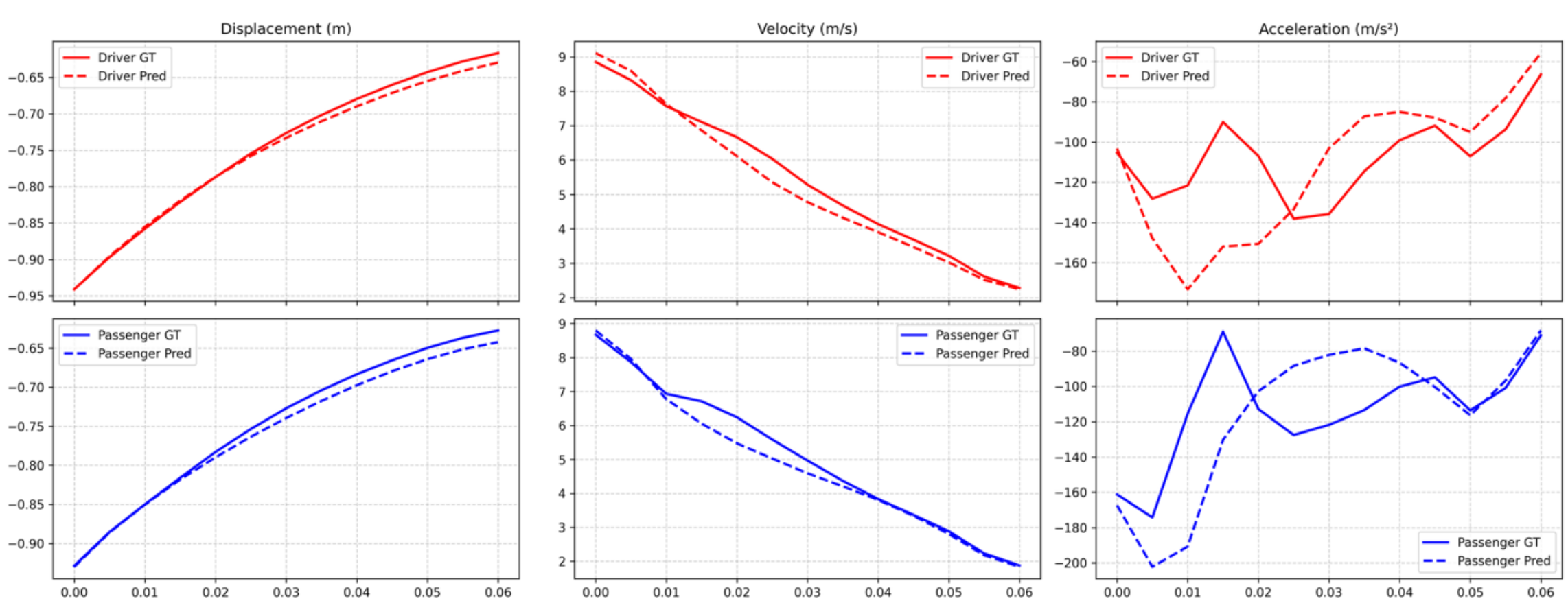

Figure 10. [Model: Transolver-AR-RT] Comparison between the predicted and ground-truth displacement, velocity, and acceleration at the driver and passenger toe pans for Sample #103 in the test dataset.

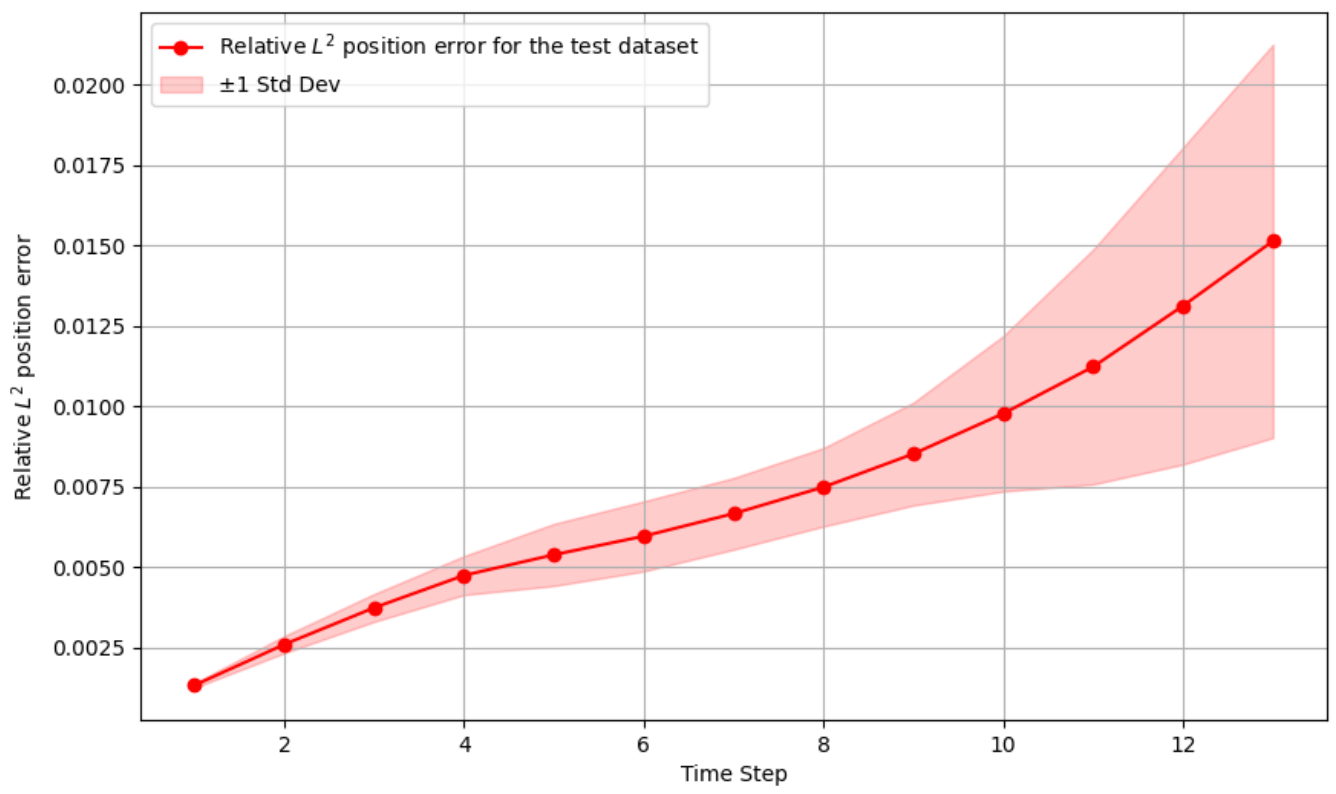

Figure 11. [Model: Transolver-AR-RT] Relative L2 position error for the test dataset.

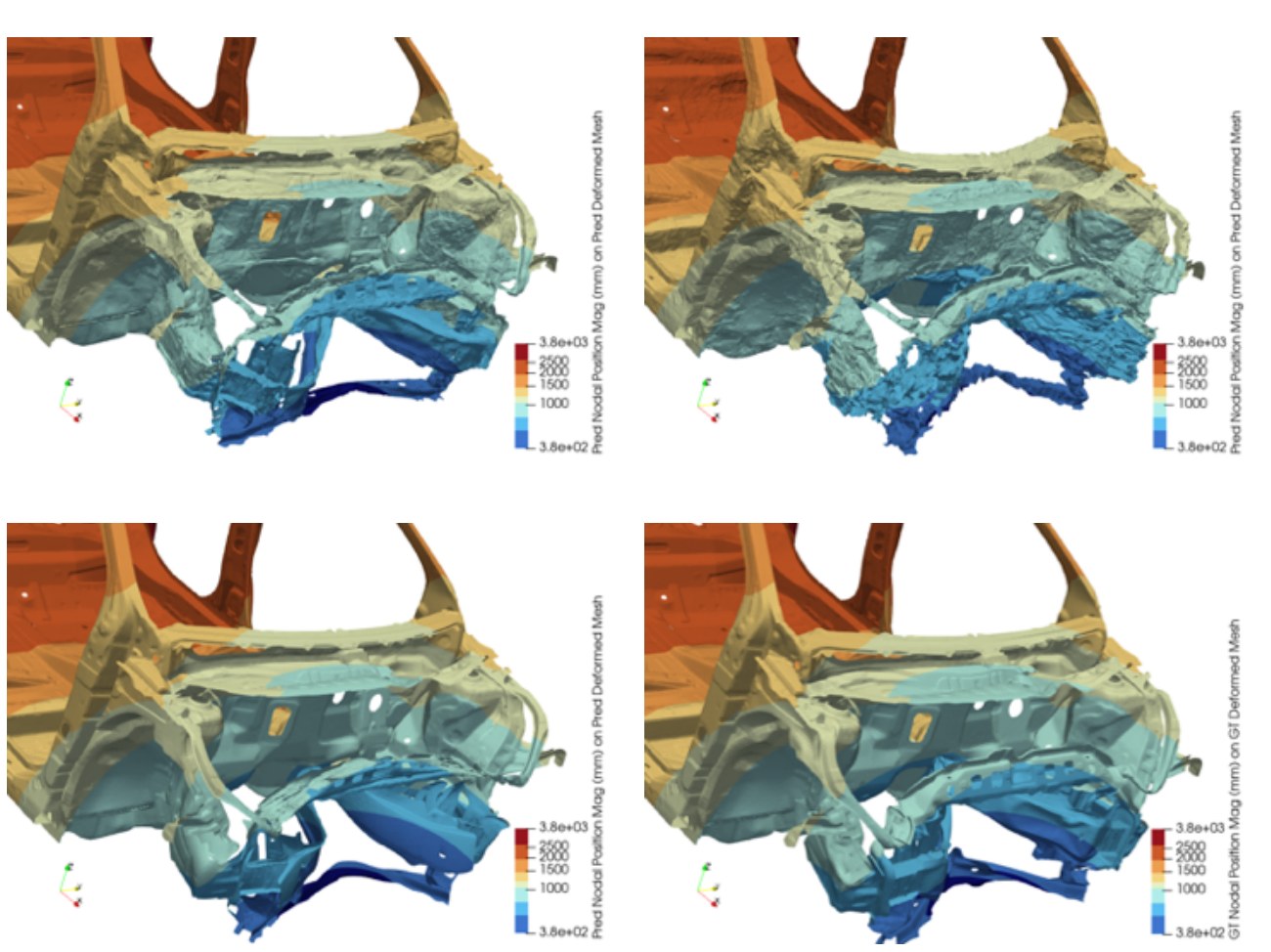

Figure 12. Comparison of results between MGN with original mesh (top left), MGN with multiscale mesh (top right) , Transolver (bottom left), and the ground truth (bottom right) for Sample #4 in the test dataset. All the models use the AR-RT scheme.

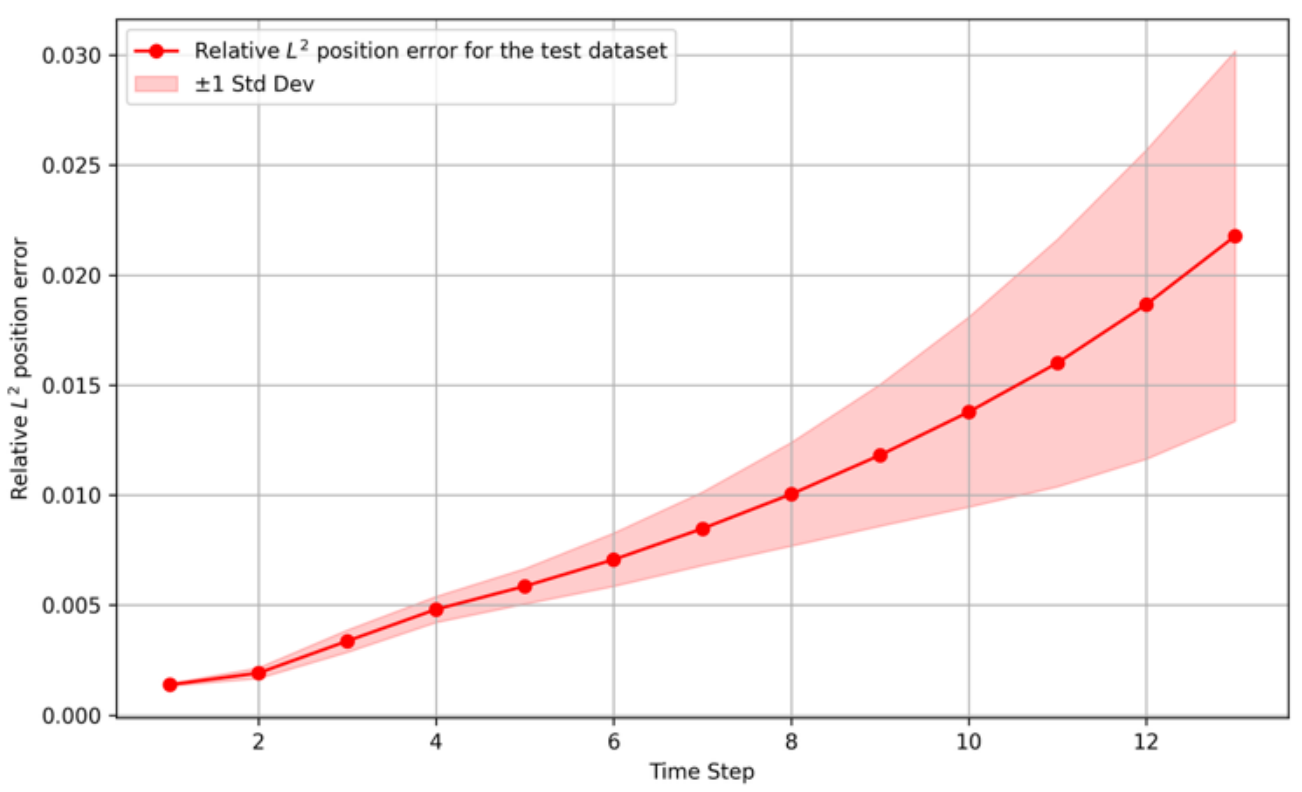

Figure 13. [Model: MGN-AR-RT, multimesh] Relative L2 position error for the test dataset. Compare this figure directly with Figure 11.

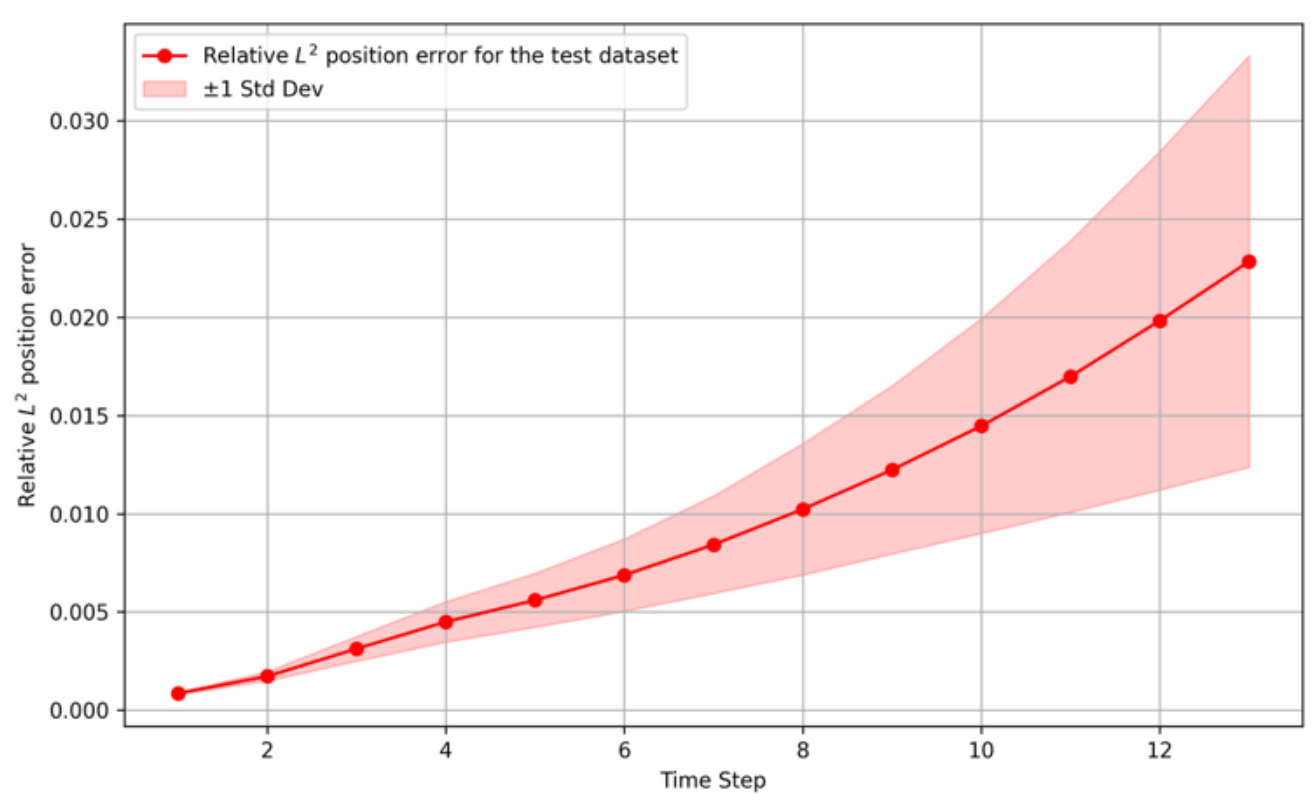

Figure 14. [Model: MGN-AR-RT, original mesh] Relative L2 position error for the test dataset. Compare this figure directly with Figures 11 and 13.

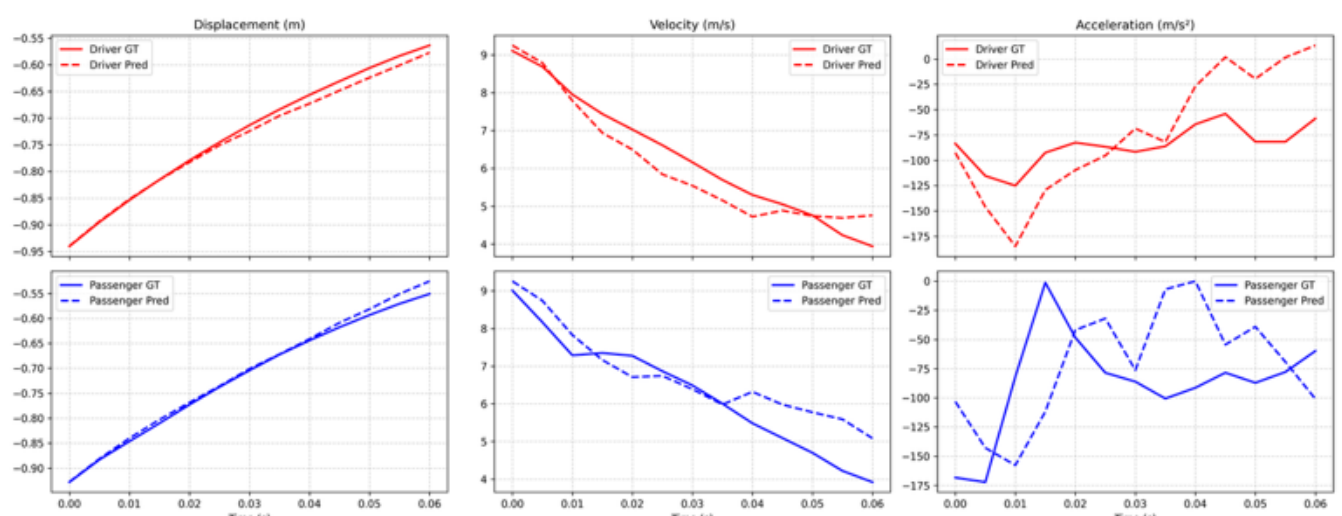

Figure 15. [Model: MGN-AR-RT, multi-scale] Comparison between the predicted and ground-truth displacement, velocity, and acceleration at the driver and passenger toe pans for Sample #4 in the test dataset. Compare this figure directly with Figure 8.

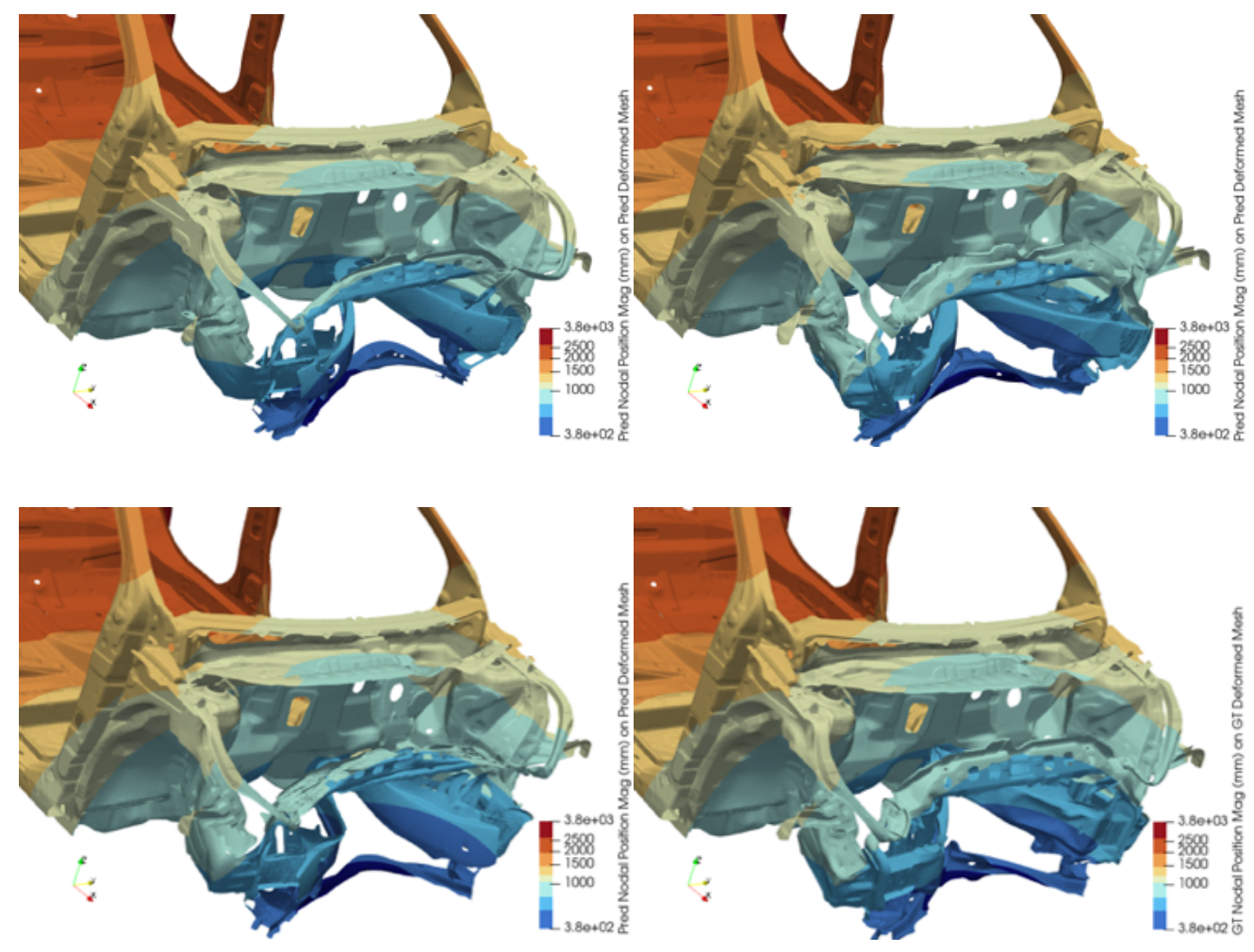

Figure 16. Comparison of Transolver results with time-conditional (top left), AR-OT (top right), and AR-RT (bottom left) schemes for Sample #4 in the test dataset. Ground truth is shown in bottom right.

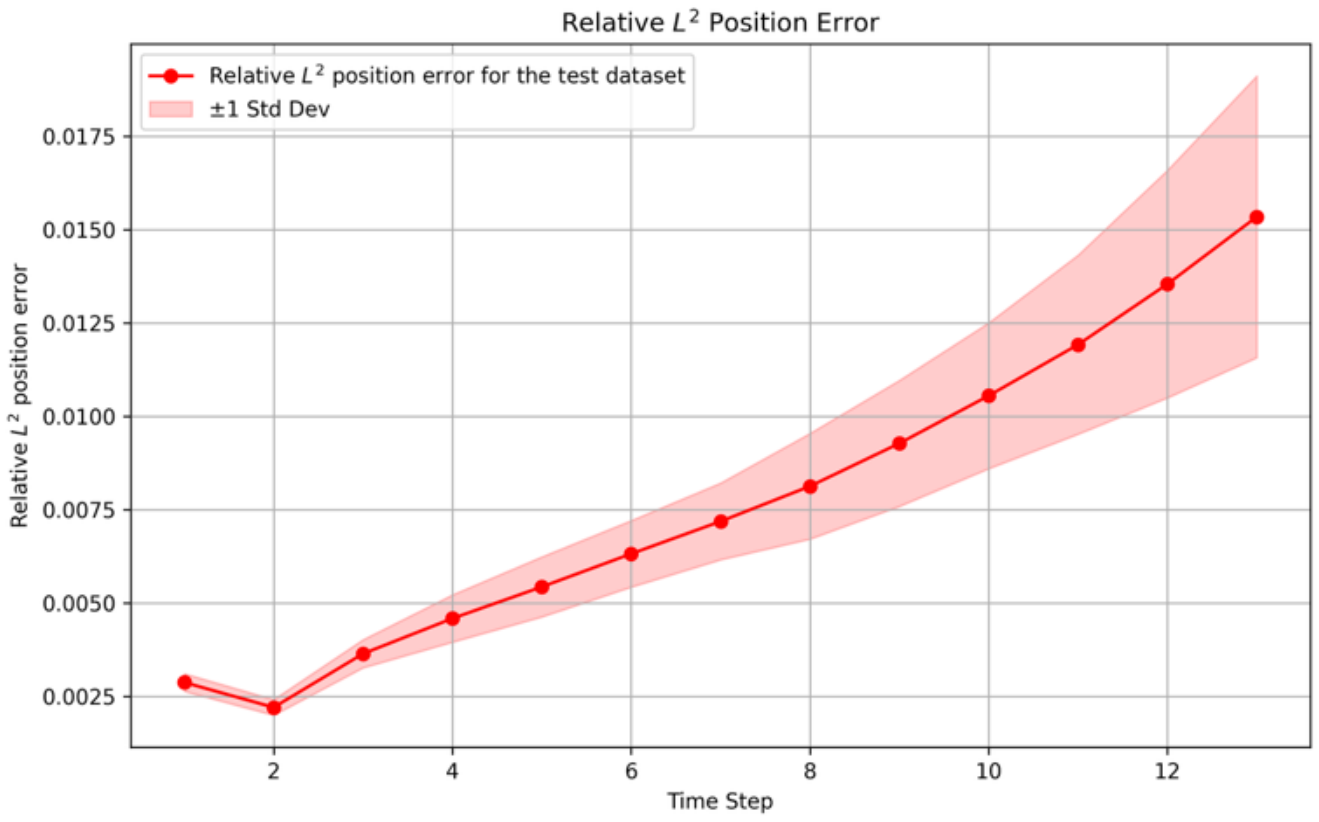

Figure 17. [Model: Time-conditional Transolver] Relative L2 position error for the test dataset. Compare this figure directly with Figure 11.

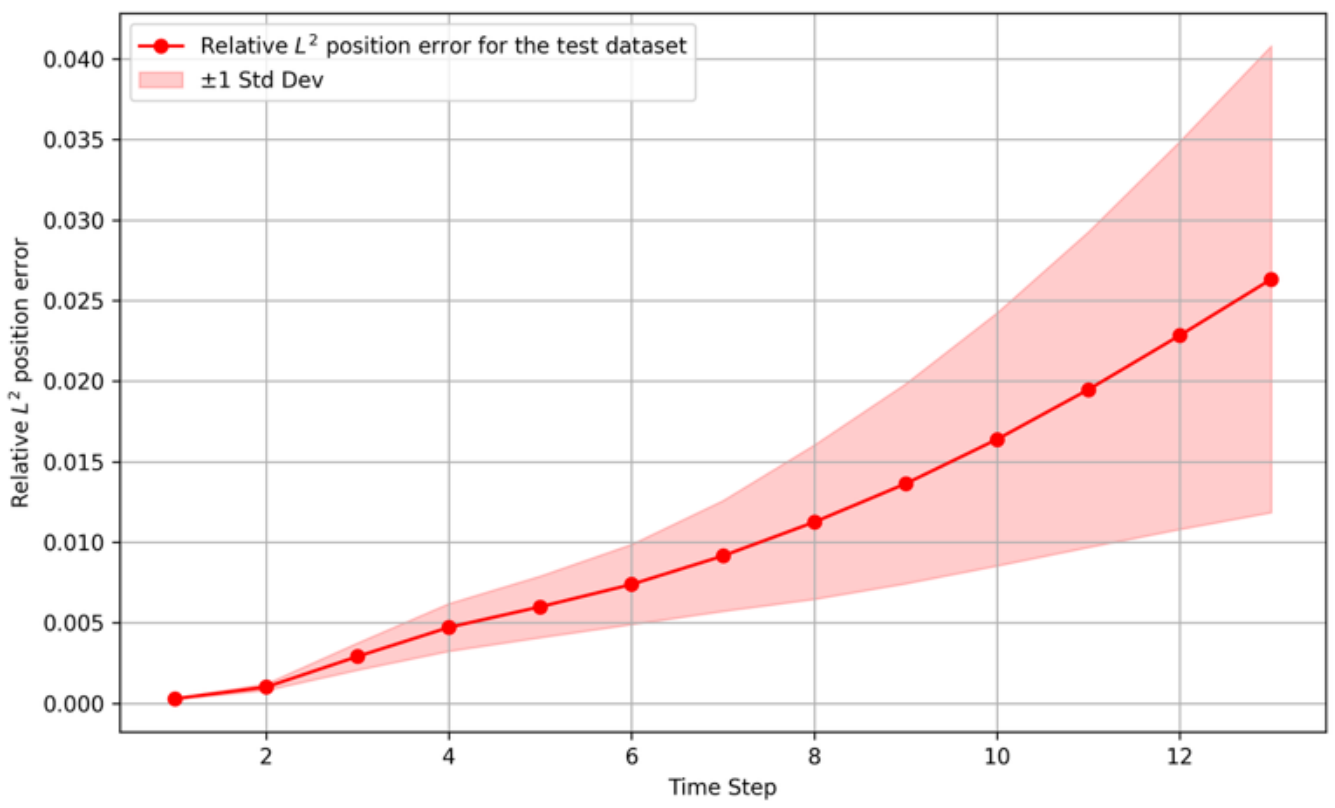

Figure 18. [Model: Transolver-AR-OT] Relative L2 position error for the test dataset. Compare this figure directly with Figure 11.

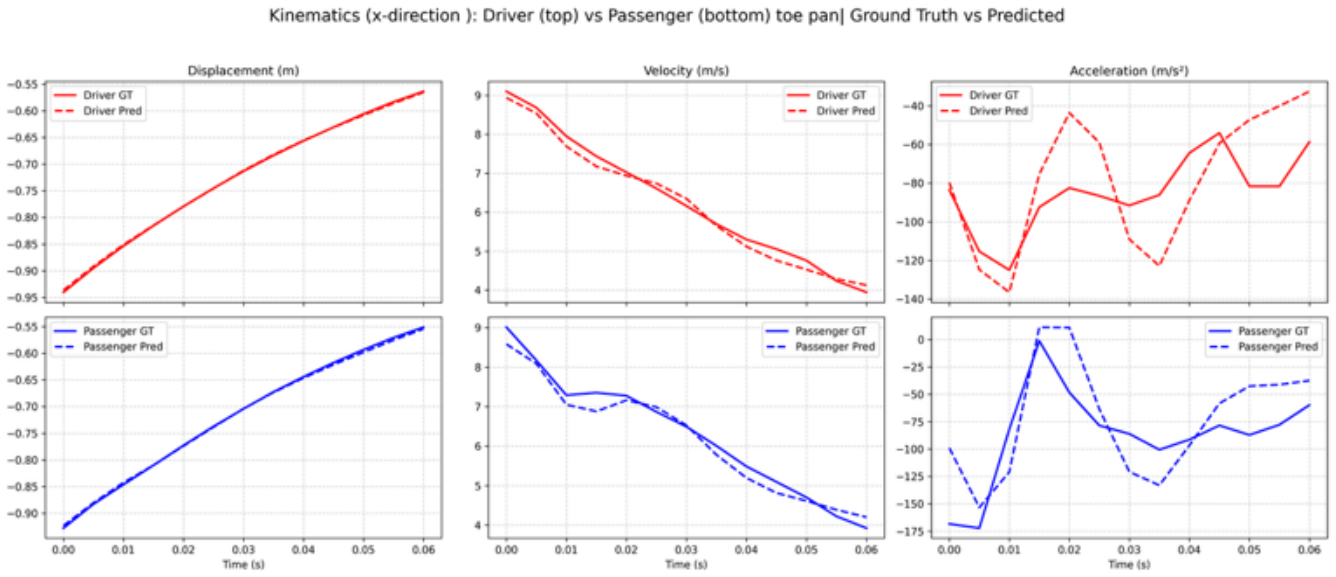

Figure 19. [Model: Time-conditional Transolver] Comparison between the predicted and ground-truth displacement, velocity, and acceleration at the driver and passenger toe pans for Sample #4 in the test dataset. Compare this figure directly with Figure 8.

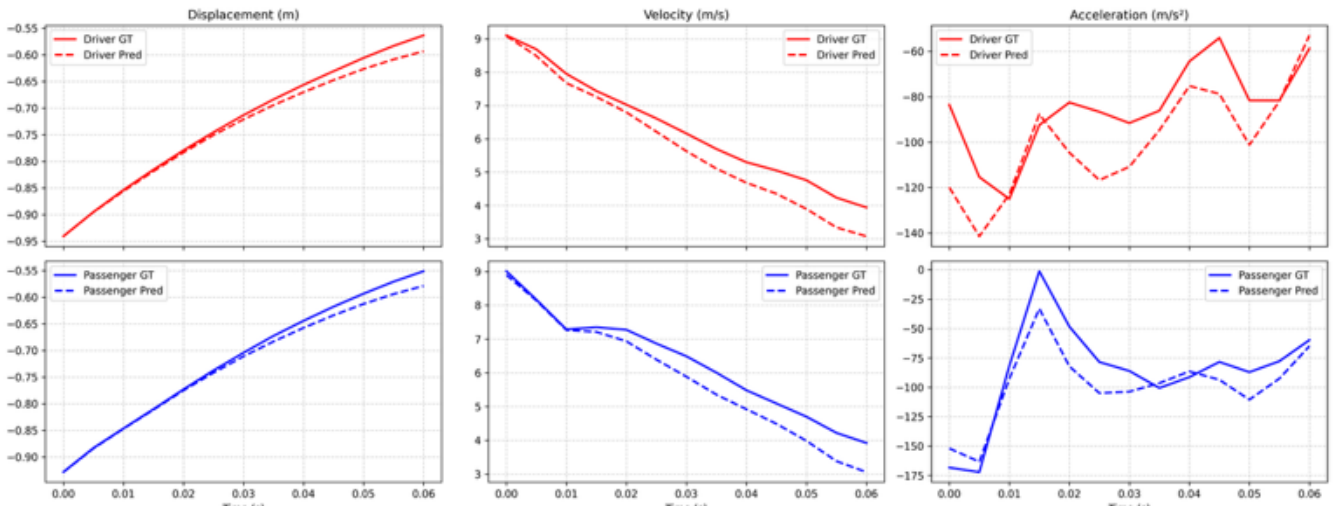

Figure 20. [Model: Transolver-AR-OT] Comparison between the predicted and ground-truth displacement, velocity, and acceleration at the driver and passenger toe pans for Sample #4 in the test dataset. Compare this figure directly with Figure 8.

The comparison of different transient modeling schemes within the Transolver framework further reveals important insights. Both the Autoregressive Rollout Training (AR-RT) and Time-Conditional schemes achieve accurate and stable predictions. The Time-Conditional model, while simpler and faster to train, treats each timestep independently and does not inherently enforce temporal causality. This limits its physical plausibility, as it lacks explicit modeling of how prior states influence subsequent deformation responses. Conversely, AR-RT explicitly enforces causality by training across multiple timesteps with backpropagation through time. The Autoregressive One-Step Training (AR-OT) approach, while offering a balance between efficiency and accuracy, exhibits faster error accumulation due to the absence of multi-step supervision.

In summary, the results confirm that both the Transolver and MeshGraphNet frameworks are viable architectures for data-driven crash dynamics modeling. Transolver achieves superior overall accuracy and stability through its latent transformer-based formulation and rollout-based temporal training, whereas Multiscale MGN provides an efficient alternative with competitive accuracy and a significantly lower computational footprint. These complementary findings underscore the potential of combining transformer-based and graph-based paradigms to develop scalable, physically consistent, and computationally efficient surrogates for structural crash simulation.

# Conclusion

This research successfully demonstrates the feasibility and effectiveness of using ML surrogate models to predict the highly non-linear, transient dynamics of automotive crash events. The framework developed in this work accurately predicts the full-field structural deformation of a complex BIW system. The predictions show strong qualitative and quantitative agreement with high-fidelity FE simulations. By replacing the iterative numerical solver of traditional FEA with a single feed-forward pass of a trained ML model, the time for a full crash simulation is reduced from minutes or hours on an HPC cluster to seconds on a single GPU workstation.

Both the Transolver and MGN architectures are promising for developing machine learning models for crash dynamics. This study explored the feasibility of using these approaches, rather than aiming to find a single "winner" model. The Transolver architecture demonstrates good predictive accuracy and stability in long-term deformation and displacement prediction. Its transformer-based latent representation enables it to aggregate global context across the spatial domain, allowing it to capture large-scale deformation patterns more cohesively than architectures that rely solely on message passing. Conversely, the MGN architecture is a competitive and interpretable alternative that excels at modeling local interactions within a mesh topology. While its predictions may show slight spatial noise compared to Transolver, the overall deformation trends remain physically consistent. The multiscale MGN variant offers an excellent balance between efficiency and accuracy, reducing training time per epoch from approximately 110 seconds to just 16 seconds while maintaining comparable accuracy to the original MGN model.

The comparison of different transient modeling schemes revealed that both Autoregressive Rollout Training (AR-RT) and the Time-Conditional scheme can achieve accurate and stable predictions. However, the AR-RT approach explicitly enforces temporal causality by training over multiple time steps with backpropagation through time. In contrast, the Time-Conditional model, which is simpler and faster to train, treats each time step independently, which limits its physical plausibility.

The immense acceleration in simulation time enabled by this framework has the potential to fundamentally change the automotive design and engineering process. Crashworthiness analysis, which is typically a late-stage validation step, can now be integrated into the earliest design phases. This allows engineers to conduct large-scale, automated design space exploration, virtually testing thousands of design variants in the time it would take to run a single FEA simulation. This capability can lead to the discovery of highly optimized designs that effectively balance safety, cost, and performance. The speed of the surrogate model also enables an interactive "digital twin" workflow, where engineers can get immediate feedback on design changes, fostering a more intuitive and creative process. Furthermore, the model can serve as a fast-running component within multi-objective optimization algorithms and generative design frameworks, paving the way for a more automated vehicle development process.

Despite these promising results, the framework has several limitations that point to important areas for future research. The model demonstrates satisfactory accuracy in predicting displacements; however, its performance in estimating velocity and acceleration at key probe points remains limited. These quantities are critical indicators of occupant safety and structural impact severity and thus cannot be overlooked. A promising direction for future work is to incorporate velocity and acceleration terms directly into the training loss, thereby encouraging the model to learn dynamic consistency and improve its predictive fidelity across all motion derivatives.

The current model effectively captures large plastic deformations but does not explicitly represent stress evolution, material fracture, element failure, or fragmentation—phenomena that are critical in realistic crash scenarios. Future work will focus on incorporating additional training data containing stress and damage information, enabling the model to learn these effects explicitly. This enhancement will enrich the model’s feature space and improve its ability to capture complex failure dynamics with higher physical fidelity. Further, extending the framework to predict these events will pose a significant challenge. It would likely require the model to learn to predict not only the state of the nodes but also to dynamically alter the graph's topology (i.e., remove edges and nodes) to represent material failure.

A limitation of the framework is its data dependency; the model's accuracy is tied to the quality and diversity of its training data. Future work could explore methods like transfer learning or the use of physics-informed neural networks (PINNs) to reduce this reliance. Another open question is the model's ability to generalize to out-of-distribution designs. While it generalizes well to unseen parameter variations, its capacity to extrapolate to radically different geometries or materials is a subject for future study.

Finally, for safety-critical applications, it is essential to provide a measure of confidence in predictions. The current model is deterministic, but future research should incorporate uncertainty quantification techniques, such as Bayesian neural networks, to provide principled confidence bounds.

This work represents a significant step towards a future where hybrid, physics-informed AI models combine the speed of deep learning with the rigor of first-principles physics, redefining the landscape of computational engineering.

# Appendix

Additional test sample results using the Transolver architecture trained with the autoregressive rollout (AR-RT) scheme are presented here.

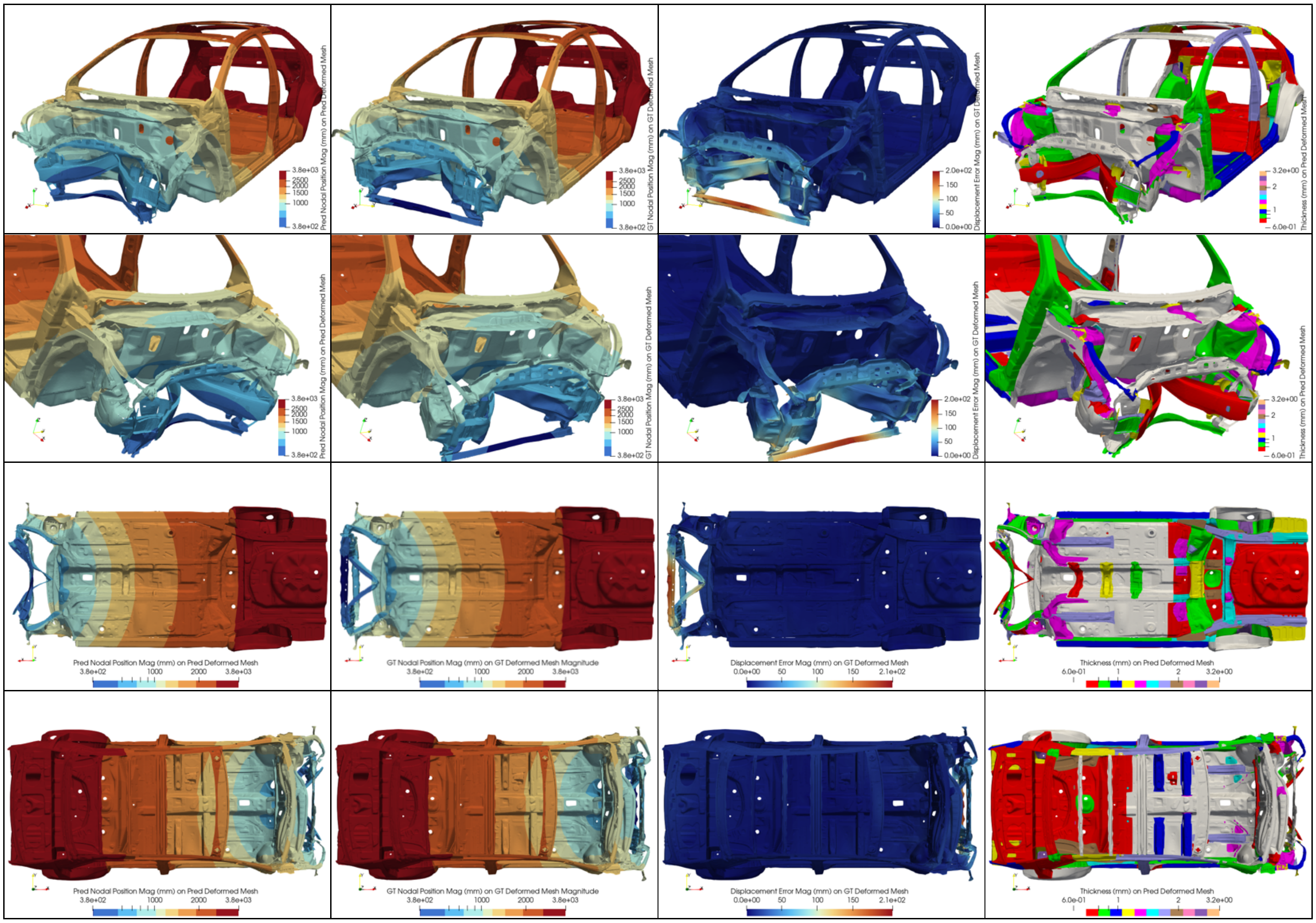

Figure A-1. [Model: Transolver-AR-RT] Visualization of crash deformation results across four viewing angles (rows) for Sample #1 in the test dataset.

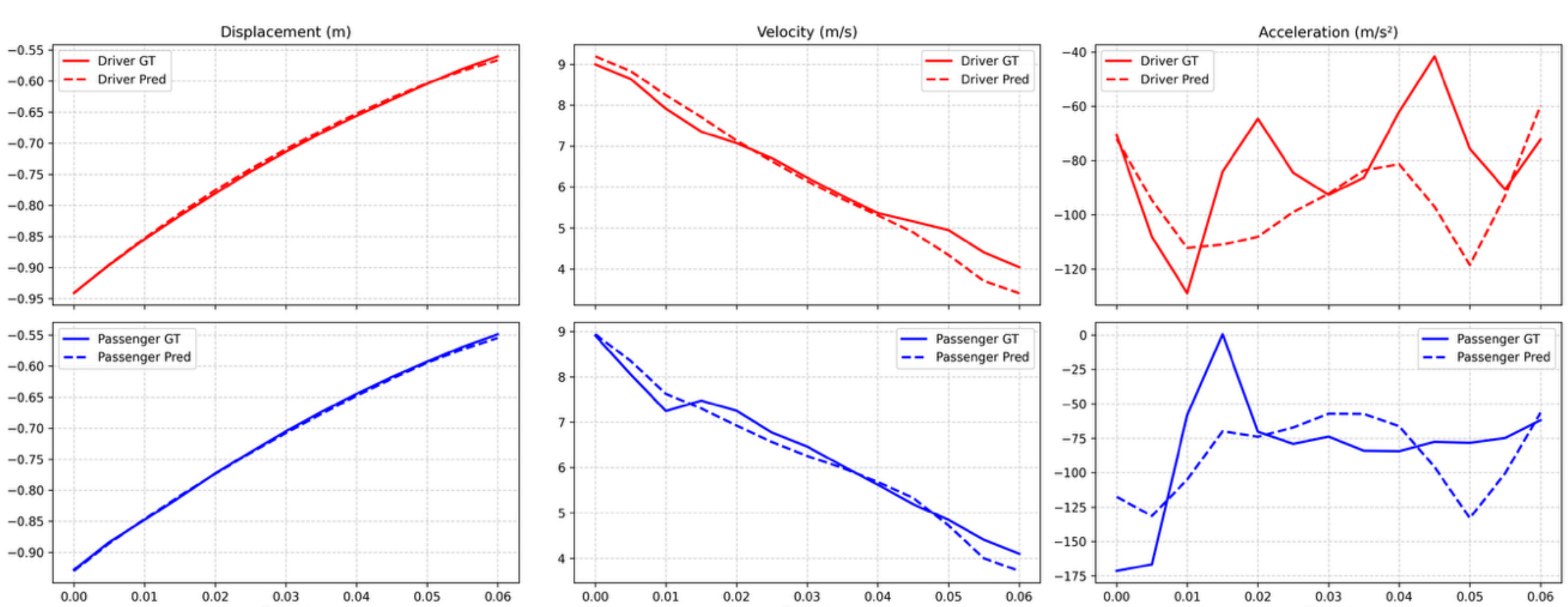

Figure A-2. [Model: Transolver-AR-RT] Comparison between predicted and ground-truth displacement, velocity, and acceleration at the driver and passenger toe pans for Sample #1 in the test dataset.

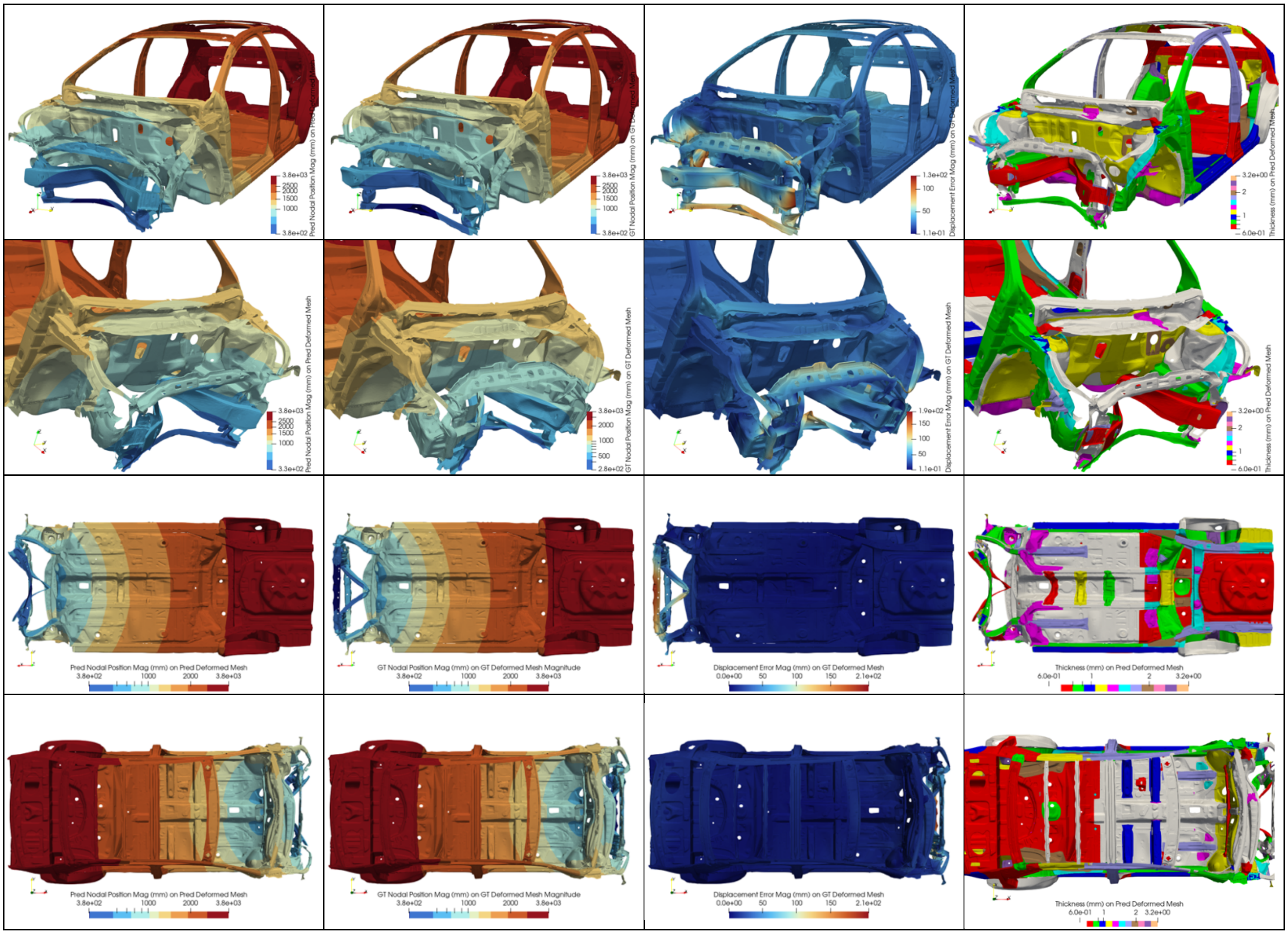

Figure A-3. [Model: Transolver-AR-RT] Visualization of crash deformation results across four viewing angles (rows) for Sample #54 in the test dataset.

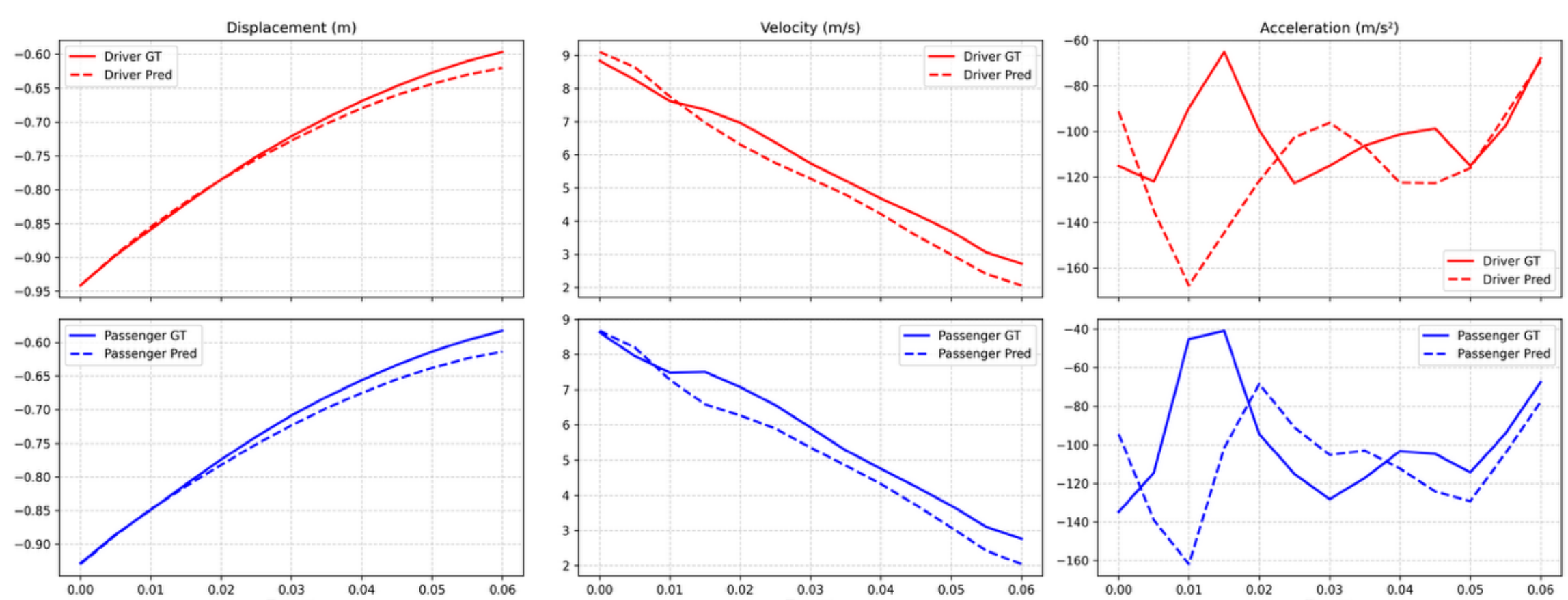

Figure A-4. [Model: Transolver-AR-RT] Comparison between predicted and ground-truth displacement, velocity, and acceleration at the driver and passenger toe pans for Sample #54 in the test dataset.